%% file: icml-submission.tex
\documentclass[nohyperref]{article}

\usepackage{microtype}
\usepackage{graphicx}
\usepackage{booktabs} 

\usepackage{subfig}
\usepackage[shortlabels]{enumitem}

\usepackage{hyperref}
\hypersetup{
    colorlinks,
    linkcolor={blue!50!black},
    citecolor={blue!50!black},
    urlcolor={blue!80!black}
}

\input{math_commands}

\newcommand{\eqdef}{\stackrel{\text{def}}{=}}

\usepackage[accepted]{icml2023}

\makeatletter

\def\paragraph{\@startsection{paragraph}{4}{\z@}{0.4ex plus
   0.4ex minus .2ex}{-0.5em}{\normalsize\bf}}
  
\makeatother

\usepackage{amsmath}
\usepackage{amssymb}
\usepackage{mathtools}
\usepackage{amsthm}

\usepackage[capitalize,noabbrev]{cleveref}

\theoremstyle{plain}
\newtheorem{theorem}{Theorem}[section]
\newtheorem{proposition}[theorem]{Proposition}

\theoremstyle{definition}

\newtheorem{assumption}[theorem]{Assumption}
\theoremstyle{remark}

\usepackage[textsize=tiny]{todonotes}

\icmltitlerunning{Better Training of GFlowNets with Local Credit and Incomplete Trajectories}

\begin{document}

\twocolumn[
\icmltitle{
Better Training of GFlowNets with Local Credit and Incomplete Trajectories 
}

\icmlsetsymbol{equal}{*}

\begin{icmlauthorlist}
\icmlauthor{Ling Pan}{milaudem}
\icmlauthor{Nikolay Malkin}{milaudem}
\icmlauthor{Dinghuai Zhang}{milaudem}
\icmlauthor{Yoshua Bengio}{milaudem,cifar}
\end{icmlauthorlist}

\icmlaffiliation{milaudem}{Mila -- Qu\'ebec AI Institute and Universit\'e de Montr\'eal}
\icmlaffiliation{cifar}{CIFAR}

\icmlcorrespondingauthor{Ling Pan}{penny.ling.pan@gmail.com}

\icmlkeywords{?}

\vskip 0.3in
]

\printAffiliationsAndNotice{} 

\begin{abstract}
Generative Flow Networks or GFlowNets are related to Monte-Carlo Markov chain methods (as they sample from a distribution specified by an energy function), reinforcement learning (as they learn a policy to sample composed objects through a sequence of steps), generative models (as they learn to represent and sample from a distribution) and amortized variational methods (as they can be used to learn to approximate and sample from an otherwise intractable posterior, given a prior and a likelihood). They are trained to generate an object $x$ through a sequence of steps with probability proportional to some reward function $R(x)$ (or $\exp(-\mathcal{E}(x))$ with $\mathcal{E}(x)$ denoting the energy function), given at the end of the generative trajectory. Like for other RL settings where the reward is only given at the end, the efficiency of training and credit assignment may suffer when those trajectories are longer. With previous GFlowNet work, no learning was possible from incomplete trajectories (lacking a terminal state and the computation of the associated reward). In this paper, we consider the case where the energy function can be applied not just to terminal states but also to intermediate states. This is for example achieved when the energy function is additive, with terms available along the trajectory. We show how to reparameterize the GFlowNet state flow function to take advantage of the partial reward already accrued at each state. This enables a training objective that can be applied to update parameters even with incomplete trajectories. Even when complete trajectories are available, being able to obtain more localized credit and gradients is found to speed up training convergence, as demonstrated across many simulations.
\end{abstract}

\section{Introduction}
Generative Flow Networks (GFlowNets)~\citep{bengio2021flow,bengio2021foundations} are variants of reinforcement learning (RL) methods~\citep{sutton2018reinforcement} trained to generate an object $x \in \mathcal{X}$ through a sequence of steps, by learning a stochastic policy $P_F$ with a training objective that would make it sample $x$ with probability proportional to a reward function $R(x)$, with sufficient capacity and training iterations.

GFlowNets are related to MCMC methods~\citep{metropolis1953equation,hastings1970monte,andrieu2003introduction} that approximately sample from a distribution associated with a given energy function or unnormalized probability function, but GFlowNets exploit the ability of machine learning (ML) to generalize and to amortize the cost of sampling (which becomes quick), at the expense of training time.
Unlike MCMC methods, they do not suffer from the mixing problem~\citep{salakhutdinov2009learning,bengio2013better,bengio2021flow}, because they do not rely on a Markov chain that makes small local steps\footnote{To mix between two modes, a Markov chain has to make the right sequences of moves to encounter a new far away and concentrated mode, and this can be very unlikely since it would require many low-probability moves.}, and instead GFlowNets generate each sample independently. On the other hand, like RL methods, they rely on exploration and the generalization power of ML to guess and discover new modes of the reward function (or regions of low energy), thanks to underlying regularities in the given energy function. 

They are also related to generative models~\citep{kingma2013auto,goodfellow2014generative,goodfellow2016deep,ho2020denoising} as they learn to represent and sample from a distribution, although they can either learn from a dataset~\citep{zhang2022generative,zhang2022unifying}, like typical deep generative models, or from an energy or reward function, like amortized variational methods~\citep{Kingma2014AutoEncodingVB,JimenezRezende2014StochasticBA} as they can be used to learn to approximate and sample from an otherwise intractable posterior, given a prior and a likelihood. As shown by~\citet{malkin2022gfnhvi}, GFlowNets can be seen as variants of amortized variational inference methods, using training objectives that are different from the usual KL objectives and enable off-policy learning and exploration methods that can avoid the factorization assumptions, importance reweighting, mode-seeking behavior and other issues with standard variational inference methods.

The sequential process for generating the composite object is related to reinforcement learning (RL) methods~\citep{sutton2018reinforcement,mnih2015human,lillicrap2015continuous,fujimoto2018addressing}.
However, because they are trained to sample proportionally to the reward rather than to maximize it, GFlowNets tend to generate a greater diversity of solutions, which can be very appealing for tasks where such diversity is desirable~\citep{jain2022biological,jain2022multi}. Note that they are related but different from entropy-regularized RL~\citep{haarnoja2017reinforcement,haarnoja2018soft}, as we will discuss in Section~\ref{sec:related_work}.

Yet, {\em previous learning objectives of GFlowNets}~\citep{bengio2021flow,bengio2021foundations,malkin2022trajectory,madan2022learning} {\em only learn from the reward of the terminal state, which is given only at the end of each trajectory}.
Due to the delayed and possibly sparse feedback from the environment (especially for the more interesting cases where the target distribution concentrates probability mass), GFlowNets may suffer from the problem of inefficient credit assignment, due to long trajectories and a highly delayed reward. This paper proposes an approach to take advantage of incomplete reward signals that are available before the terminal state is reached, even allowing to benefit from incomplete trajectories, that do not reach a terminal state, i.e., without knowledge of the terminal reward.

As a motivating example, consider the possibility of using GFlowNets to sample abstract compositional objects akin to thoughts, as previously suggested~\citep{bengio2021foundations,gfn-tutorial2022}.
The human brain's working memory can only hold around a handful of symbols at a time, and although a sequence of such memory contents can correspond to a probabilistic inference \cite{baddeley1992working, cowan1999embedded} (e.g., an interpretation for an image or a video in terms of latent variables that can explain it), this working memory bottleneck -- see the work on the global workspace theory~\citep{baars1993cognitive,dehaene1998neuronal,shanahan2005applying,shanahan2006cognitive,shanahan2010embodiment,shanahan2012brain,dehaene2017consciousness} -- generally prevents us from forming a complete description of the full explanation of the input. This suggests that the brain can learn from partial inferences that do not contain a full specification of the latent variables that explain the given input.

However, variational methods and current GFlowNet training objectives~\citep{bengio2021flow,malkin2022trajectory,madan2022learning,bengio2021foundations} require complete specification of the latent explanation (i.e., a complete trajectory in the case of GFlowNets) before receiving a reward (such as how well the generated explanation fits the observed data and the prior). 
To perform the kind of computation associated with working memory and forming sequences of thoughts, GFlowNets would need to be modified to accommodate training trajectories that do not reach a terminal state. 
How could that even be possible if no reward is available before a terminal state  is reached, i.e., before a full specification of the latent explanation is constructed? 

In many cases, the reward function can actually be decomposed into a product of factors, where some of these factors can be accrued along the trajectory. A natural case is where the reward 
$R(x)=\exp(-\mathcal{E}(x))$ corresponds to an energy function $\mathcal{E}(x)=\sum_t \mathcal{E}_t$ that is additive with terms that can be obtained along the trajectory (e.g., at transitions $t$ of the trajectory), as in GFlowNets over sets~\citep{bengio2021foundations}. In the case of a sparse factor graph that models the joint of latents and observed variables, each time the variables associated to a clique of the graph are specified, we can compute an energy term~\citep{bengio2021foundations}.

In this paper, we focus on the underlying technical question: how to train GFlowNets with incomplete trajectories? We find that the proposed solution also accelerates training even when complete trajectories are available. We propose Forward-Looking GFlowNets (FL-GFN), a novel formulation that exploits the ability to compute an energy value (even if incomplete) for intermediate states, in order to deliver a local credit signal and gradient as soon as a local reward factor is accrued. 
The code is publicly available at \url{https://github.com/ling-pan/FL-GFN}.

The main contributions are summarized as follows:
\begin{itemize}
\item We propose a novel GFlowNets formulation, FL-GFN, that can exploit the existence of a per-state or per-transition energy function.
\item We show that with FL-GFN, we can still guarantee convergence, to fit and match the distribution corresponding to the given reward function, while credit information is available early. We show how FL-GFN can be trained from incomplete trajectories without access to terminal states.
\item We conduct extensive experiments on the set generation and bit sequence generation tasks, which demonstrates the effectiveness of FL-GFN. It is also found to be scalable to the more complex and challenging molecule discovery task.
\end{itemize}

\section{Background}
\subsection{GFlowNet Preliminaries}

Let $G = (\mathcal{S}, \mathbb{A})$ be a directed acyclic graph (DAG), where $\mathcal{S}$ is the set of vertices (states), and $\mathbb{A}\subset\gS\times\gS$ is the set of edges (called actions). A unique {\em initial state} $s_0 \in \mathcal{S}$ is defined, without incoming edges. States without outgoing edges are called \emph{terminal},\footnote{The alternative convention of \citet{bengio2021foundations} defines terminal states as those with an outgoing edge to a designated sink state $s_f$. The two conventions are equivalent, as noted by \citet{malkin2022trajectory}, but ours has the advantage of allowing simpler notation for the expressions needed in this paper.} and the set of terminal states is denoted by $\gX \subset \mathcal{S}$. A sequence $\tau=(s_0\rightarrow s_1\rightarrow\dots\rightarrow s_n)$, with $s_n\in\gX$ and $(s_i\rightarrow s_{i+1})\in\mathbb{A}$ for all $i$, is called a \emph{complete trajectory}.

Let $R:\gX\to\R_{\geq0}$ be a nonnegative reward function on the set of terminal states. The goal of GFlowNets is to learn a stochastic policy $P_F$, specified as a distribution over the children of every nonterminal state in $G$, such that complete trajectories starting at $s_0$ and taking actions sampled from $P_F$ terminate at $x \in \mathcal{X}$ with likelihood proportional to the reward $R(x)$. That is, if $P_F^\top(x)$ is the marginal likelihood that a complete trajectory $s_0\rightarrow s_1\rightarrow\dots\rightarrow s_n$ sampled from $P_F$ has $s_n=x$, then we desire $P_F^\top(x)\propto R(x)$. Note that $P_F^\top(x)$ is a sum of likelihoods of all trajectories leading to $x$:
\[P_F^\top(x)\eqdef\sum_{\tau=(s_0\rightarrow\dots\rightarrow s_n=x)}P_F(\tau)=\sum_{\tau}\prod_{i=1}^nP_F(s_i|s_{i-1}),\]
which may be an intractably large sum. The GFlowNet forward policy $P_F$ constructs objects sequentially, by modifying the partial object at each timestep, and transitions sampled from $P_F$ should be thought of as incremental construction actions: unlike in standard RL, $G$ has no cycles, which means that the same state cannot be visited twice within a trajectory. This can easily be achieved by including a time stamp in the state~\citep{bengio2021foundations}.

The action policy $P_F$ -- parameterized as a neural network $P_F(s_i|s_{i-1}; \theta)$ taking a state $s_{i-1}$ as input and producing a distribution over the children $s_i$ of $s_{i-1}$ -- is the main object of GFlowNet training. 

\subsection{Training Criteria for GFlowNets} \label{sec:gfn_bg}

There are several auxiliary quantities that can be either introduced in the training process or computed after a policy has been trained, depending on the training objective chosen. We summarize three objectives that are relevant in this paper as follows:

\paragraph{Detailed balance (DB).} Two auxiliary quantities are learned: a scalar \emph{state flow function} $F(s;\theta)$ and a \emph{backward policy} $P_B(s|s';\theta)$, which is a distribution over the parents $s$ of every noninitial state $s'$. We set $F(x)=R(x)$ for terminal states $x$. The DB constraint (that will only be approximately achieved, thanks to a training objective) enforces that for any action $s\rightarrow s'$, we have
\begin{equation}
F(s;\theta)P_F(s'|s;\theta)=F(s';\theta)P_B(s|s';\theta).
\label{eq:db_constraint}
\end{equation}
For a given forward policy $P_F$, there is a unique backward policy $P_B$ and state flow function $F$ that satisfy Eq.~(\ref{eq:db_constraint}) \citep{bengio2021foundations}. 
In practice, the constraint is enforced by taking gradient steps with respect to $\theta$ on the square of the log-ratio between the two sides of Eq.~(\ref{eq:db_constraint}), where the edge $s\to s'$ is chosen from a training policy. To improve exploration and prevent the agent from getting trapped around a few modes of $R$~\citep{jain2022biological,pan2022gafn}, the training policy is usually chosen as a tempered version of the forward policy or its mixture with a uniform policy, i.e., $\pi_{\theta} = \epsilon U + (1 - \epsilon) P_F$ that resembles $\epsilon$-greedy exploration in RL.
\citet{bengio2021flow} prove that if the training policy $\pi_{\theta}$ has full support and the expected loss is minimized globally, then $P_F^\top(x)\propto R(x)$.

\paragraph{Trajectory balance (TB).} With this GFlowNet objective, the learned auxiliary quantities are a backward policy $P_B$, as above, and a scalar $Z_\theta$ (typically parametrized in the log-domain as $\log Z_\theta$). For any complete trajectory $\tau=(s_0\rightarrow s_1\rightarrow\dots\rightarrow s_n=x$, the TB constraint is
\begin{equation}
Z_\theta\prod_{i=1}^nP_F(s_i|s_{i-1};\theta)=R(x)\prod_{i=1}^nP_B(s_{i-1}|s_i;\theta).
\label{eq:tb_constraint}
\end{equation}
Satisfaction of this constraint for all complete trajectories also implies that $P_F^\top(x)\propto R(x)$~\citep{malkin2022trajectory}. The constraint can again be enforced by optimizing the squared log-ratio between the left and right hand sides of Eq.
~(\ref{eq:tb_constraint}).

\paragraph{Subtrajectory balance (SubTB).} The parametrization is the same as in DB. The subtrajectory balance (SubTB) constraint applies to partial trajectories $s_m\rightarrow\dots\rightarrow s_n$, where $s_m$ and $s_n$ are not necessarily initial and final:
\begin{align}
&F(s_m;\theta)\prod_{i=m+1}^nP_F(s_i|s_{i-1};\theta)\nonumber\\=&F(s_n;\theta)\prod_{i=m+1}^nP_B(s_{i-1}|s_i;\theta).
\label{eq:subtb_constraint}
\end{align}
Special cases for this constraint are DB (when the partial trajectory has length $1$) and TB (when the trajectory is complete, noting that $F(x;\theta)=R(x)$ when $x$ is terminal). Satisfaction of the SubTB constraint for all partial trajectories of a given length does not necessarily imply sampling proportionally to the reward, but its satisfaction for partial trajectories of all lengths (thus including complete trajectories and terminal states) implies both the DB and TB constraints hold and is thus sufficient to guarantee $P_F^\top(x)\propto R(x)$. \citet{madan2022learning} empirically tested the training objective based on the SubTB constraint, which reduces the gradient variance of the TB objective.

\section{Related Work} \label{sec:related_work}
\paragraph{GFlowNets.} There have been many recent efforts in applying GFlowNets in a number of settings, e.g., molecule discovery~\citep{bengio2021flow}, sequence design~\citep{jain2022biological}, Bayesian structure learning~\citep{deleu2022bayesian, nishikawa2022bayesian}, and providing theoretical understandings of GFlowNets~\citep{bengio2021foundations,malkin2022gfnhvi,zimmermann2022variational,zhang2022unifying,lahlou2023continuousgfn}. 
Given its practical importance, several studies also emerged~\citep{bengio2021foundations,malkin2022gfnhvi,madan2022learning} to improve the learning efficiency of GFlowNets since the proposal of the flow matching learning objective initially proposed by~\citet{bengio2021flow}.
It can also be jointly trained with an energy or reward function~\citep{zhang2022generative}.
\citet{pan2022gafn} introduce intrinsic exploration rewards into GFlowNets in an additive way to help exploration in sparse reward tasks.
There have also been recent efforts in extending GFlowNets to stochastic environments with stochasticity in transition dynamics~\citep{pan2023stochastic} and rewards~\citep{zhang2023distributional}.
Yet, GFlowNets could suffer from inefficient credit assignment since they only learn from a trajectory reward provided at terminal states, which poses a critical challenge to the attribution of credit on each of the actions in a trajectory, especially those in the early parts of the trajectory.
Previous learning objectives of GFlowNets require access to terminal states and need to be trained with complete trajectories, which can be infeasible when the composite terminal states $x$ have considerably large sizes.

\paragraph{Reinforcement Learning (RL).} RL agents typically aim to learn a reward-maximizing policy, instead of learning to sample in proportion to the reward function.
Soft Q-learning~\citep{haarnoja2017reinforcement} introduces entropy regularization in its objective~\citep{haarnoja2018soft,Zhang2022LatentSM} and learns a stochastic energy-based policy. However, it can perform badly in general (non-tree) DAGs as shown in~\citet{bengio2021flow}, since there can be a potentially large number of trajectories leading to the same terminal state, and some terminal states (corresponding to longer trajectories) may thus have exponentially more trajectories into them, which biases learning in favor of longer trajectories.
In episodic RL settings such is the case with GFlowNets, the agent only receives a trajectory feedback at the end of each trajectory, which can hinder learning efficiency in long-horizon problems~\citep{ren2021learning}. 

\begin{figure*}[!h]
\centering
\subfloat[]{\includegraphics[width=0.4\linewidth]{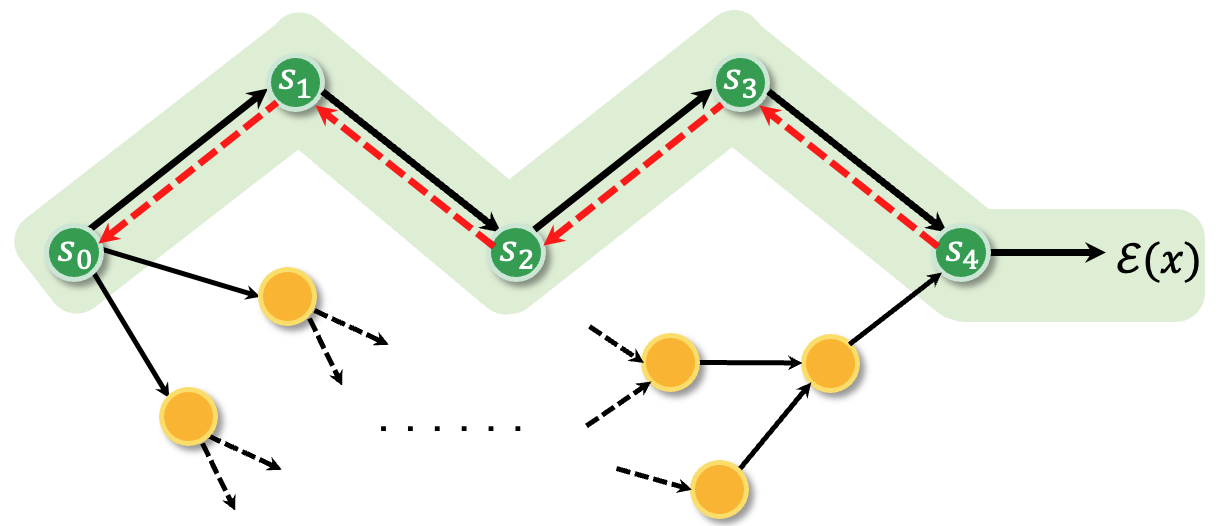}} 
\hspace{.3in}
\subfloat[]{\includegraphics[width=0.4\linewidth]{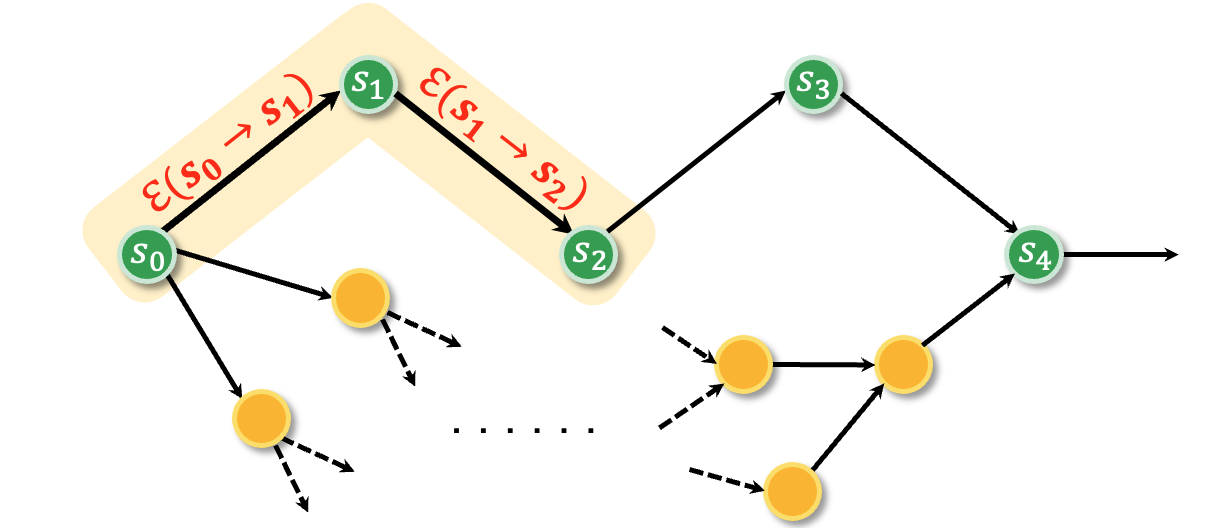}}
\caption{(a) Reward propagation in previous learning objectives of GFlowNets and (b) in FL-GFN, with additive energy terms for each transition. Note that (a) requires learning based on complete trajectories since it only learn from the terminal energy $\mathcal{E}(x)$, while (b) makes it possible to learn from short incomplete trajectories.}
\label{fig:reward_prpagate}
\end{figure*}

\section{Proposed Method} \label{sec:method}
Previous GFlowNets learning objectives, e.g., detailed balance (DB)~\citep{bengio2021foundations}, trajectory balance (TB)~\citep{malkin2022trajectory}, and sub-trajectory balance (SubTB)~\citep{madan2022learning} therefore require learning with complete trajectories.
This is because they only learn from the energy of the terminal state, i.e., $\mathcal{E}(x)$, and therefore need to visit terminal states $x$ to obtain informative learning signals, as illustrated in Figure~\ref{fig:reward_prpagate}(a).
The requirement of learning from complete trajectories is a critical limitation when the trajectory length is long or the final composite object is complex, as motivated from the above working memory bottleneck~\citep{baars1993cognitive} example and the situation of inferring the latent explanation for a rich input such as a complex image or a video~\citep{gfn-tutorial2022}.
In addition, learning from a highly delayed reward also makes it challenging for agents to properly associate their actions to future rewards, and thus hinders the efficiency of learning and credit assignment~\citep{ren2021learning}.

In this section, we introduce our approach, Forward-looking GFlowNets or FL-GFN for short to take advantage of the computability of energy at intermediate states, or equivalently, an additive energy decomposition.
FL-GFN can take intermediate reward signals into account, and thus obtains faster learning and enables learning from incomplete trajectories. We also theoretically justify the proposed learning objective and discuss its connections to existing approaches.
FL-GFN relies on the following assumption:
\begin{assumption}
\label{as:general-energy}
The terminal state energy function ${\cal E}:\cal X \rightarrow \R$ can be extended to the set of all states, i.e., there exists ${\cal E}:\cal S \rightarrow \R$.
\end{assumption}
As a consequence, we can also define an energy function over transitions: 
\begin{equation}
{\cal E}(s\rightarrow s') \eqdef {\cal E}(s') - {\cal E}(s).
\end{equation}
Similarly, we can define the energy differential associated with all trajectories from $s$ to $x$ as
\begin{equation}
{\cal E}(s\rightarrow x) \eqdef {\cal E}(x) - {\cal E}(s).
\end{equation}
We obtain that the energy of a state $s_t$ (terminal or not) can be written additively over transitions for any trajectory $s_0 \rightarrow s_1 \rightarrow \dots \rightarrow s_n$ if we define ${\cal E}(s_0)\eqdef 0$:
\begin{equation}
{\cal E}(s_t) = \sum_{i=1}^n {\cal E}(s_{i-1}\rightarrow s_i).
\end{equation}

\subsection{Sources of Partial Energies}

We now describe two common settings where energies for incomplete states may arise. 

\paragraph{Purely additive energies.} We can start from a given per-transition energy function ${\cal E}(s\rightarrow s')$ and apply the above to define a per-state energy function. For example, this applies to the construction of a set whose log-reward is a sum of terms depending on its elements (cf.~the Set-GFN construction in \citet{bengio2021foundations}, where a state corresponds to a set of elements, a transition corresponds to inserting a new element in the current set, and an energy term is associated with each element in $s$). Another scenario is the sampling of posteriors over latent variables with compositional structure \citep{gfn-em}, where our proposed methodology is applied to parse trees under probabilistic context-free grammars.

\paragraph{State spaces extending the target space.} A second motivating example is the case where all states belong to the same space (e.g., a vector space or a space of structured objects that can be processed by a neural network) and the energy can be calculated for any state $s$ using the same function that is applied in the computation of the reward for terminal states $x$. In this case, the energy $\mathcal{E}(s\rightarrow s')$ measures the stepwise marginal improvement in the log-reward. Note that this formulation applies even if $s$ and $s'$ are not complete objects from which termination is allowed, as long as an extension of the reward function to nonterminal states can be evaluated (e.g., the molecule generation domain as studied in Section~\ref{sec:mol}). 

\subsection{Forward-looking GFlowNets (FL-GFN)}

Consider the flow at $s$ and exploit Assumption~\ref{as:general-energy} to rewrite the terminal energy ${\cal E}(x)$ as ${\cal E}(s) + {\cal E}(s\rightarrow x)$: 
\begin{align} 
F(s) &= \sum_{x\geq s} P_B(s|x) e^{-{\cal E}(x)} \\
   &= e^{-{\cal E}(s)} \sum_{x\geq s} P_B(s|x) e^{{-{\cal E}(s\rightarrow x)}}. 
\end{align}
Here $P_B(s|x)$ denotes the probability of reaching the intermediate state $s$ from the terminal state $x$ via the one-step backward policy $P_B(s|s')$.
The idea is to take advantage of the fact that we already know ${\cal E}(s)$ when we have reached state $s$ and that we can thus factor it out of the right-hand side, as above. With this in mind, we define
the {\bf forward-looking flow} as
\begin{equation}
\widetilde{F}(s) \eqdef e^{{\cal E}(s)} F(s) = \sum_{x\geq s} P_B(s|x) e^{{-{\cal E}(s\rightarrow x)}},
\label{eq:fl_flow}
\end{equation}
i.e., $\widetilde{F}(s)$ only contains a sum over terms with energies of transitions to come (hence the name of forward-looking flow). 
Using this, we can write a variant of the detailed balance constraint (Eq.~(\ref{eq:db_constraint})), called the FL-DB constraint, expressed in terms of forward-looking flows:
\begin{equation} 
\widetilde{F}(s) P_F(s' | s) = \widetilde{F}(s') P_B(s | s') e^{{-\mathcal{E}(s \to s')}}. 
\label{eq:fmconstraint_fldb}
\end{equation}
Note that this constraint can be extended in the same way that DB was extended to SubTB (as we will study in Section~\ref{sec:subtb}).
Based on the resulting dense supervision (where actual energy/reward signals are available at intermediate steps), we can hypothesize that more efficient credit assignment is achieved. 
In addition, it also makes it possible for GFlowNets to learn based on incomplete trajectories without access to terminal states, so long as the training sequences of intermediate energies over incomplete trajectories contain enough information to generalize over the energies to be expected downstream. 

\paragraph{Implementation.}

In practice, we train the FL-GFN based on the FL-DB constraint in Eq.~(\ref{eq:fmconstraint_fldb}), following Algorithm~\ref{alg}, to minimize the corresponding loss function over transitions $s\rightarrow s'$ in the log-space:
\begin{equation} 
\begin{split}
\mathcal{L}(s, s') = & \left(\log \widetilde{F}(s; \theta) + \log P_F(s' | s; \theta) \right.\\
& \left.- \log \widetilde{F}(s'; \theta) - \log P_B(s | s'; \theta)  + {\mathcal{E}(s \to s')}\right)^2. 
\end{split}
\label{eq:loss_fldb}
\end{equation}

An alternative implementation is to make the following substitution and optimize for the DB constraint Eq.~(\ref{eq:db_constraint}):
\begin{equation} \label{eq:reparam}
    \log F(s)=-\gE(s)+\log\tilde F(s;\theta),
\end{equation}
where $\log\tilde F(s;\theta)$ is a learned model.

\begin{algorithm}[t]
\caption{Training Step of FL-GFN}
\begin{algorithmic}[1]
\STATE Initialize forward and backward policies $P_{F}$, $P_{B}$, and the energy-to-go flow $\tilde{F}$ with parameters $\theta$\\ 
\FOR {each transition $s\rightarrow s'$ sampled from a training trajectory}
\STATE Measure the transition energy $\mathcal{E}(s \to s' )$ \\
\STATE Update $\theta$ by $\theta - \eta \nabla_{\theta} \mathcal{L}(s_t, s_{t+1})$ as per Eq.~(\ref{eq:loss_fldb})\\
\ENDFOR
\end{algorithmic}
\label{alg}
\end{algorithm}

\paragraph{Theoretical justification.} We now theoretically justify that if we reach a global minimum of the expected value of the FL-GFN loss (Eq.~(\ref{eq:loss_fldb})), then the FL-GFN samples from the target reward distribution correctly. The proof can be found in Appendix~\ref{app:proof}.
\begin{proposition}
Suppose that Assumption~\ref{as:general-energy} is satisfied, which makes it possible to define a forward-looking flow as per Eq.~(\ref{eq:fl_flow}). If $\mathcal{L}(s, s') = 0$ for all transitions, then the forward policy $P_F(s'|s; \theta)$ samples proportionally to the reward function.
\end{proposition}

Next, we discuss its connection with existing methods in the GFlowNets literature.
\paragraph{Connection with existing methods.}
The forward-looking flow $\widetilde{F}(s;\theta)$ in Eq.~(\ref{eq:fl_flow}) can also be interpreted as a new parametrization of the state flow function as in Eq.~(\ref{eq:reparam}).
The case of ${\cal E}(s)\equiv0$ for non-terminal states $s$ corresponds to regular DB (or SubTB) training, in which learning signals come only from the reward at the end of a trajectory and the flow function is learned directly as a neural network, since $F(s)=\widetilde{F}(s)$.
However, in many cases, there is a natural ${\cal E}(s)$ available as discussed above and shown in the experiments below.

If termination is permitted from any state $s$ (into terminal state $s^\top$) with reward $e^{-{\cal E}(s)}$ and the GFlowNet constraints are satisfied, we have $F(s)P_F(s^\top|s)=R(s^\top)=e^{-{\cal E}(s)}$. Substituting Eq.~(\ref{eq:reparam}), this simplifies to $\log\widetilde{F}(s;\theta)=-\log P_F(s^\top|s)$. The loss introduced by \citet{deleu2022bayesian} 
can be considered as the DB loss with the parametrization in Eq.~(\ref{eq:reparam}) and directly setting $\log\widetilde{F}(s;\theta)=-\log P_F(s^\top|s)$.

\section{Experiments}
We conducted extensive experiments to investigate the following key questions and understand the effectiveness of the proposed approach:
i) How does the forward-looking approach compare against previous GFlowNet learning objectives in terms of learning efficiency and final performance? 
ii) Can it learn given incomplete trajectories only when it does not have access to terminal states? 
iii) Can it be applied to different GFlowNet learning objectives? 
iv) How does it work on more complex and challenging tasks?

\subsection{Set Generation}
\subsubsection{Experimental Setup}
We first conducted a series of experiments on a didactic set generation task with set GFlowNets~\citep{bengio2021foundations} to understand the effect of the forward-looking approach. 
The agent generates a set of size $|S|$ from $|U|$ distinct elements sequentially.
At each timestep, the agent chooses to add an element from $U$ to the current set $s$ (the GFlowNet state) without repeating elements, and gets an energy term for adding the element (a fixed value for each element).
The task terminates when there are exactly $|S|$ elements in the set $s$, and the total energy for constructing $s$ is $\mathcal{E}(x) = \sum_{t \in \tau}\mathcal{E}(t)$, where $\tau$ is the sampled trajectory and $t \in \tau$ is a transition.
We consider different scales of the set generation task including small, medium, and large with increasing sizes of $|S|$ and $|U|$.
More details of the experimental setup can be found in Appendix~\ref{app:set_setup}.

\subsubsection{Performance Comparison}
\begin{figure}[!ht]
\centering
\subfloat[]{\includegraphics[width=0.5\linewidth]{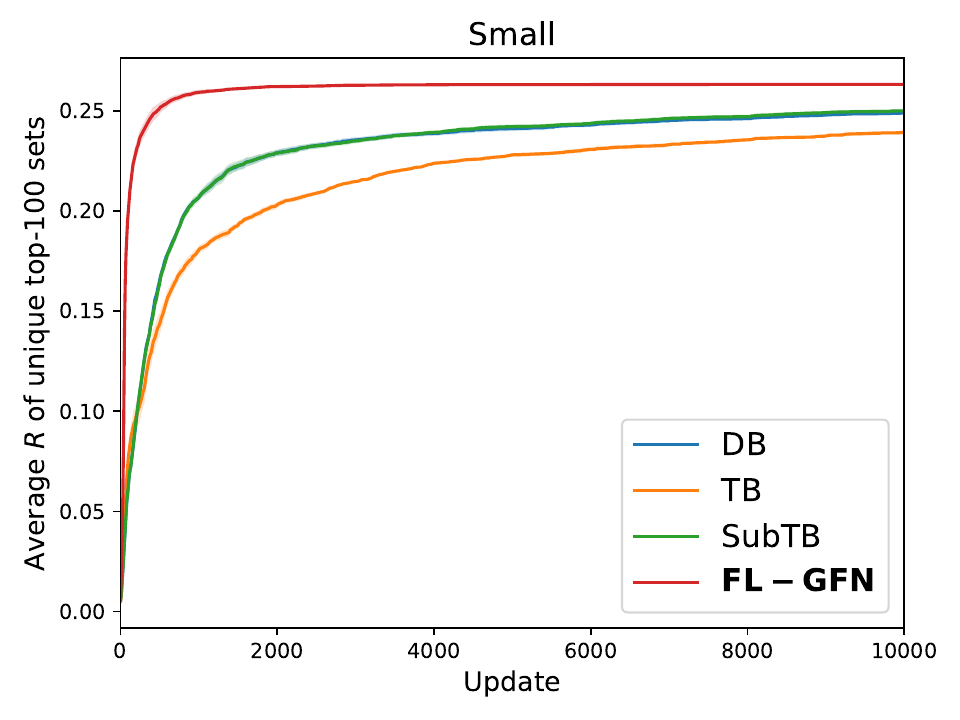}}
\subfloat[]{\includegraphics[width=0.5\linewidth]{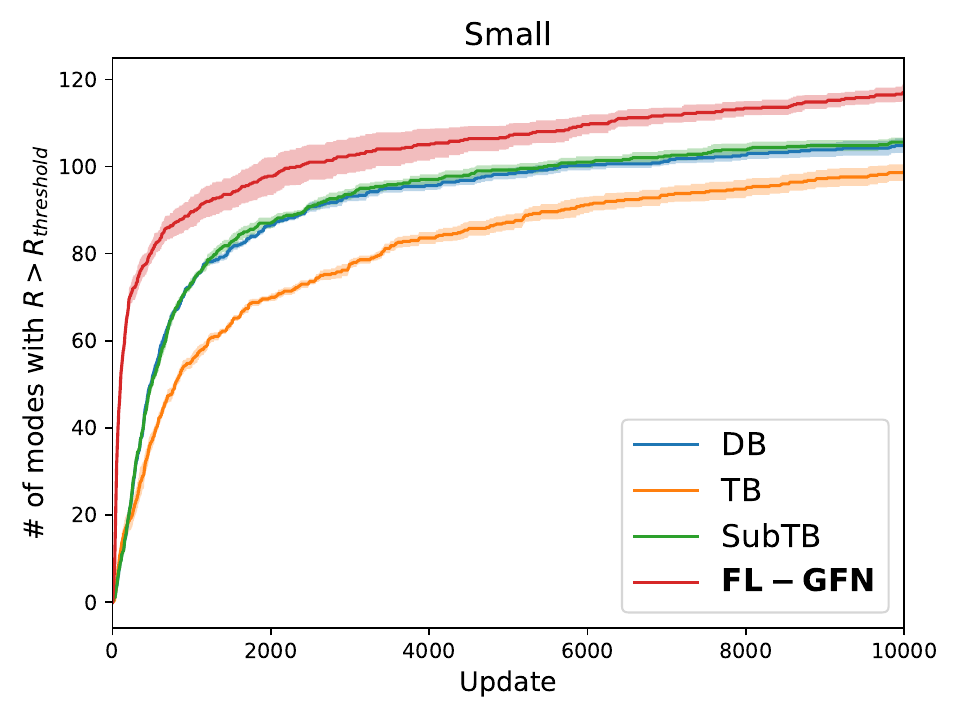}}\\
\subfloat[]{\includegraphics[width=0.5\linewidth]{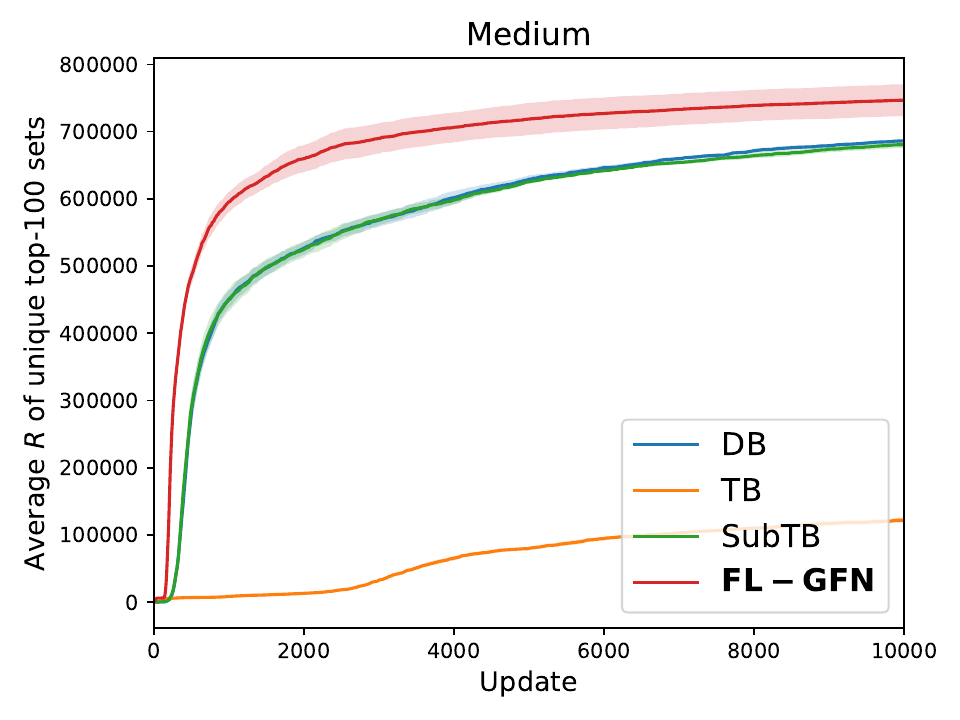}} 
\subfloat[]{\includegraphics[width=0.5\linewidth]{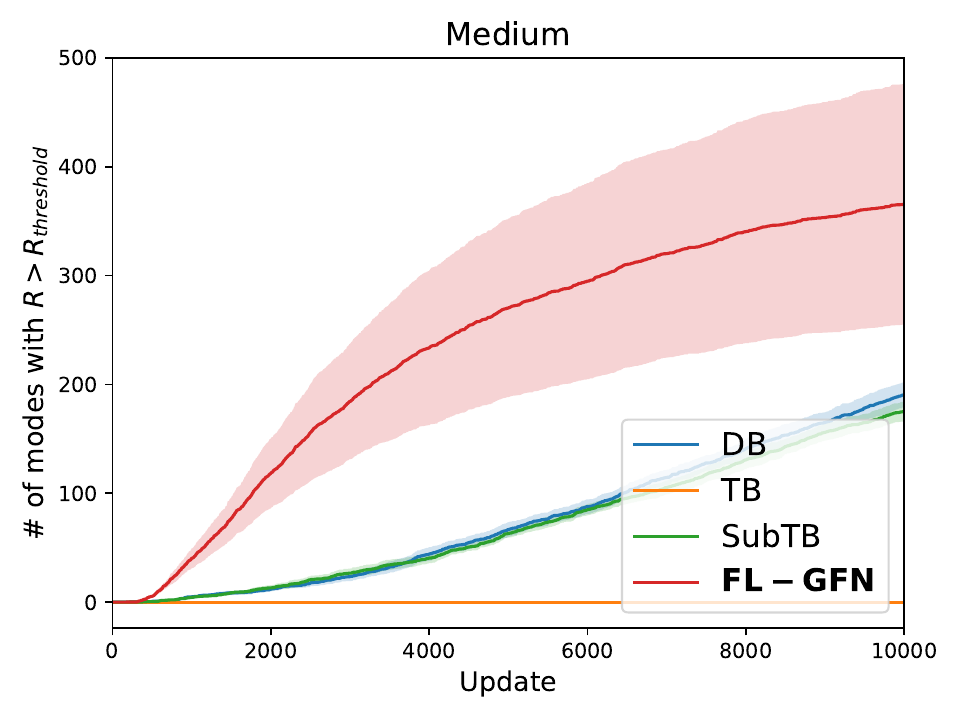}} \\
\subfloat[]{\includegraphics[width=0.5\linewidth]{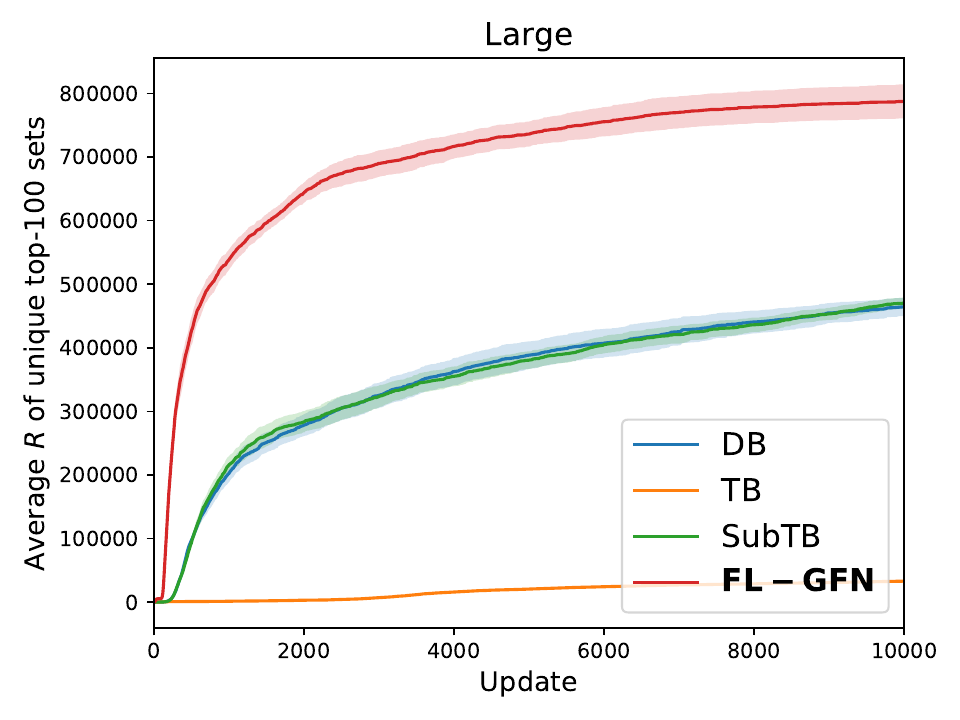}}
\subfloat[]{\includegraphics[width=0.5\linewidth]{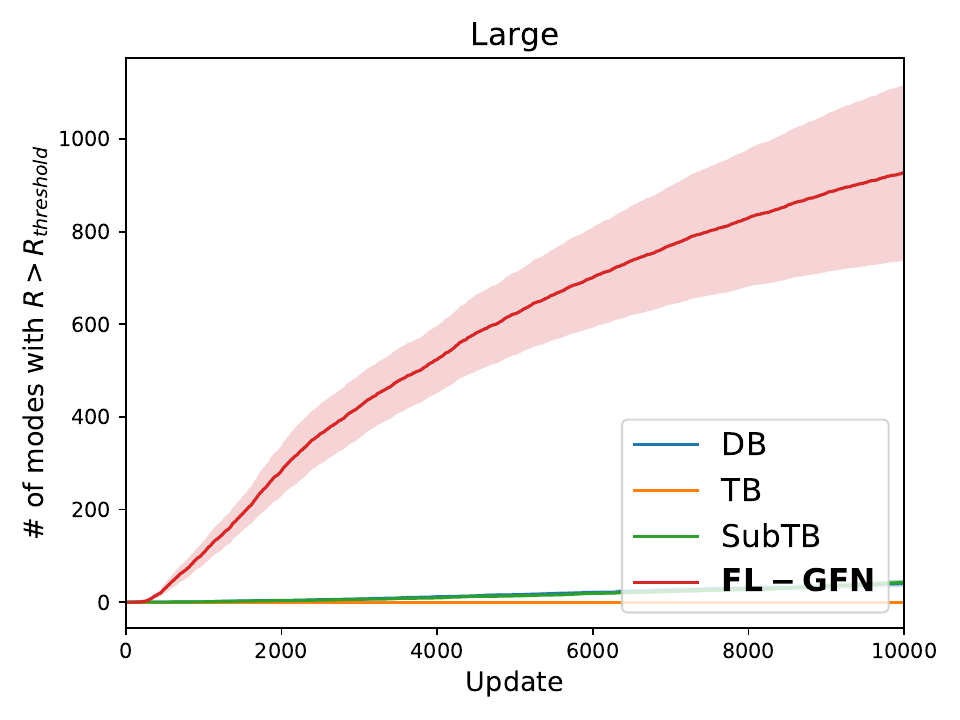}}
\caption{Comparison on the set generation task with (a, b) small, (c, d) medium, and (c, d) large sets.
The first column shows the advantage of FL-GFN in terms of the average reward of the unique top-$100$ sets, the second column in terms of the number of modes discovered by each method, i.e., diversity.}
\vspace{-.1in}
\label{fig:set_perf}
\end{figure}

\begin{figure*}[!h]
\centering
\subfloat[]{\includegraphics[width=0.28\linewidth]{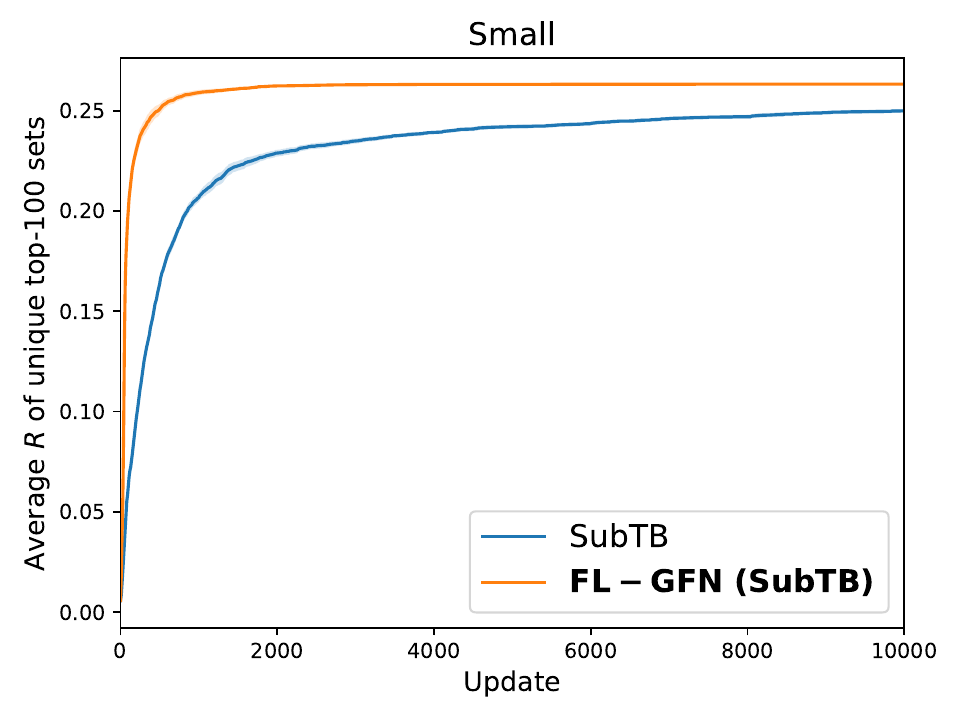}}
\subfloat[]{\includegraphics[width=0.28\linewidth]{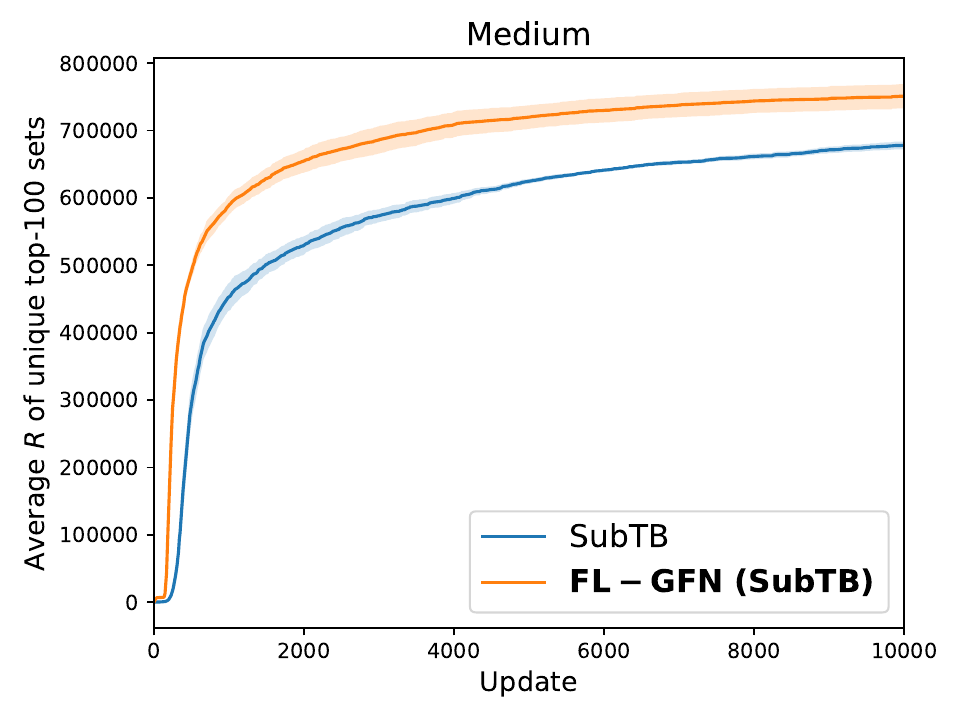}}
\subfloat[]{\includegraphics[width=0.28\linewidth]{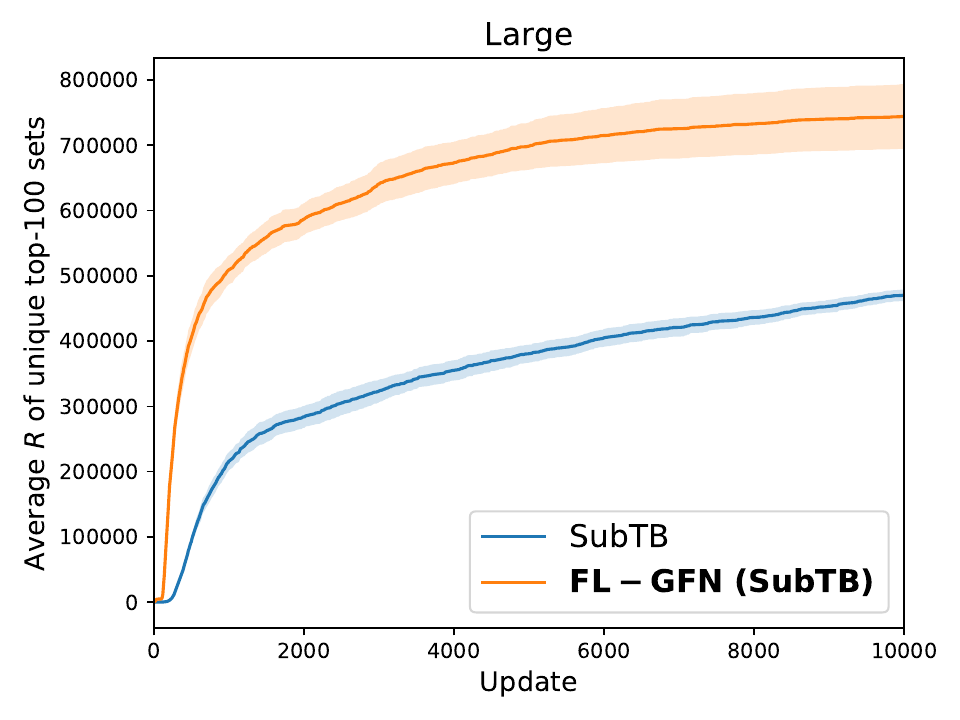}}
\caption{Performance comparison of FL-GFN (SubTB) and SubTB in the set generation task with different problem sizes including (a) small (b) medium (c) large. FL-GFN (SubTB)converges faster and to better solutions, especially for larger sets.}
\label{fig:set_subtb}
\end{figure*}

\begin{figure*}[!h]
\centering
\subfloat[]{\includegraphics[width=0.28\linewidth]{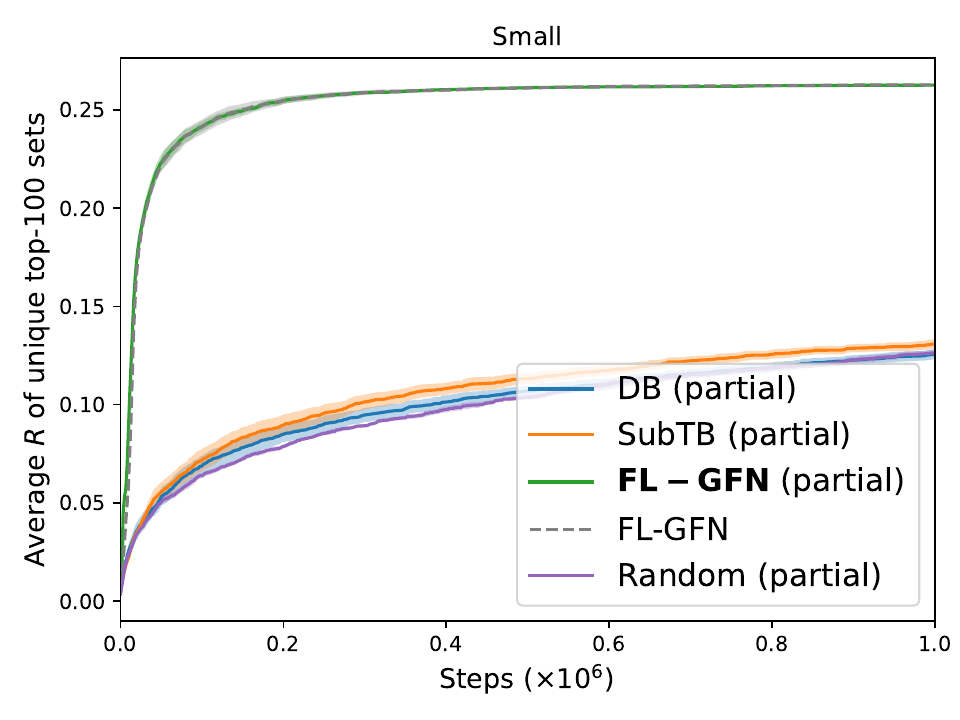}}
\subfloat[]{\includegraphics[width=0.28\linewidth]{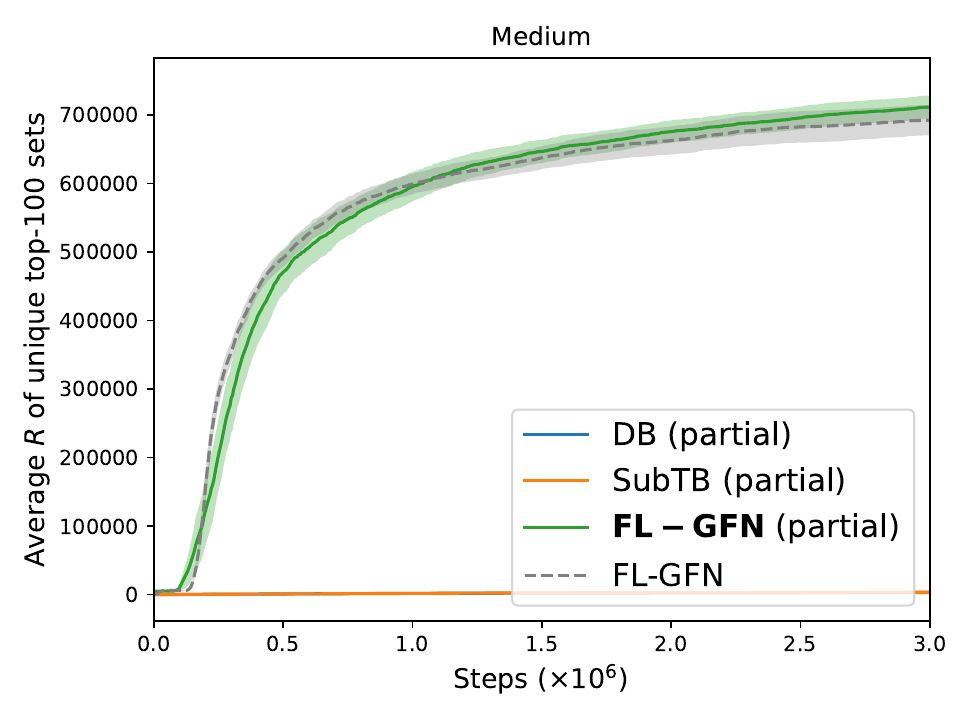}}
\subfloat[]{\includegraphics[width=0.28\linewidth]{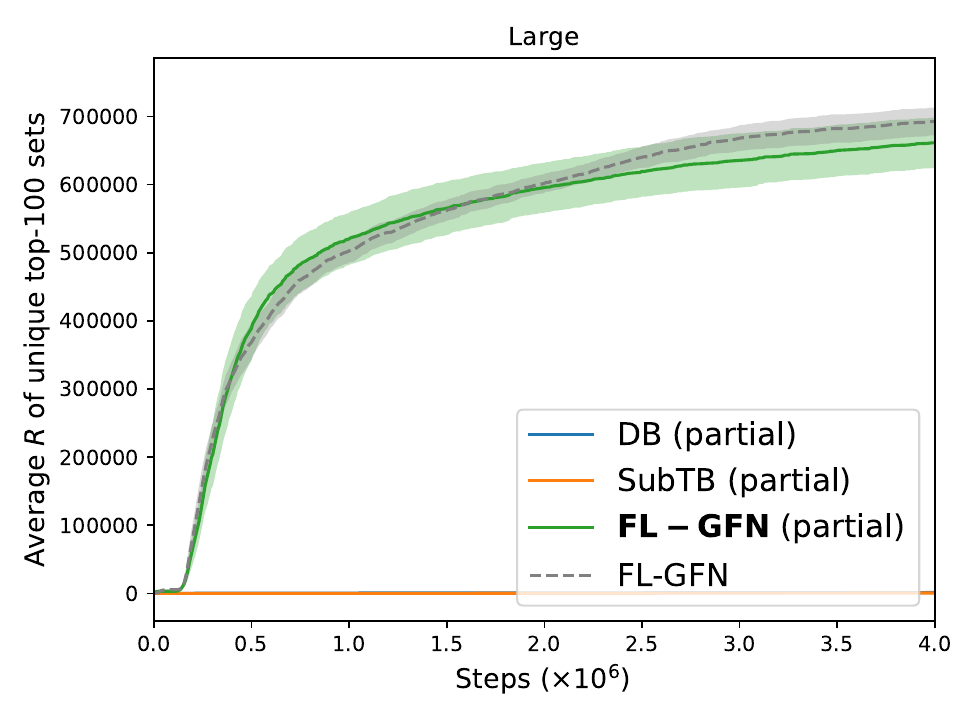}}
\caption{Performance comparison of baselines trained with incomplete trajectories in the set generation task with different scales including (a) small (b) medium and (c) large. The advantage of FL-GFN increases with the length of trajectories.}
\label{fig:set_partial}
\end{figure*}

Following~\citet{bengio2021flow}, we evaluate the methods in terms of the average reward of the unique top-$100$ sampled candidates (from all the samples generated during training) and the number of modes discovered by each algorithm, so as to measure both performance and diversity.
We compare FL-GFNs with previous GFlowNets learning objectives including detailed balance (DB), trajectory balance (TB), and subtrajectory balance (SubTB).

The first column in Figure~\ref{fig:set_perf} demonstrates the quality of generated sets in the set generation task with different problem sizes including small, medium, and large in each row.
As shown, the forward-looking approach significantly outperforms previous baselines including DB, TB, and SubTB in training efficiency and quality of the solutions, with a more significant gap in larger-scale environments, presumably because of FL-GFN's more efficient credit assignment mechanism.
We also observe that the performance of TB degrades with longer trajectories in the larger-scale set generation tasks, presumably due to the larger variance~\citep{madan2022learning}, and SubTB performs closely to DB.
The second column in Figure~\ref{fig:set_perf} illustrates the number of modes with rewards above a threshold discovered by each method, diversity of the good solutions, and FL-GFN discovers many more high-reward modes faster.

\subsubsection{Applicability to Other Objectives} \label{sec:subtb}
FL-GFN is versatile and can be applied to different learning objectives of GFlowNets except for TB, so long as Assumption ~\ref{as:general-energy} is satisfied. 
Here, we investigate the applicability of FL-GFN with the SubTB constraint~\citep{madan2022learning}, yielding the following FL-SubTB constraint (and the corresponding training objective) as in Eq.~(\ref{eq:fl_subtb})
following Section~\ref{sec:method} and the constraint of SubTB in Eq.~(\ref{eq:subtb_constraint}), where $\widetilde{F}$ denotes the forward-looking flow, and $\mathcal{E}(s \to s')$ denotes the intermediate transition energy.
\vspace{-.1in}
\begin{equation}
\begin{split}
& \widetilde{F}(s_i; \theta) \prod_{t=i}^{j-1} P_F(s_{t+1} | s_t; \theta) \\
= &\widetilde{F}(s_j; \theta) \prod_{t=i}^{j-1} P_B(s_t | s_{t+1}; \theta) \prod_{t=i}^{j-1}e^{-\mathcal{E}(s_t \to s_{t+1})}
\label{eq:fl_subtb}
\end{split}
\end{equation}

As demonstrated in Figure~\ref{fig:set_subtb}, FL-GFN (SubTB) consistently outperforms SubTB with an improved average reward of unique top-$100$ sets.

\begin{figure*}[t]
\vspace*{-2mm}
\centering
\subfloat[]{\includegraphics[width=0.28\linewidth]{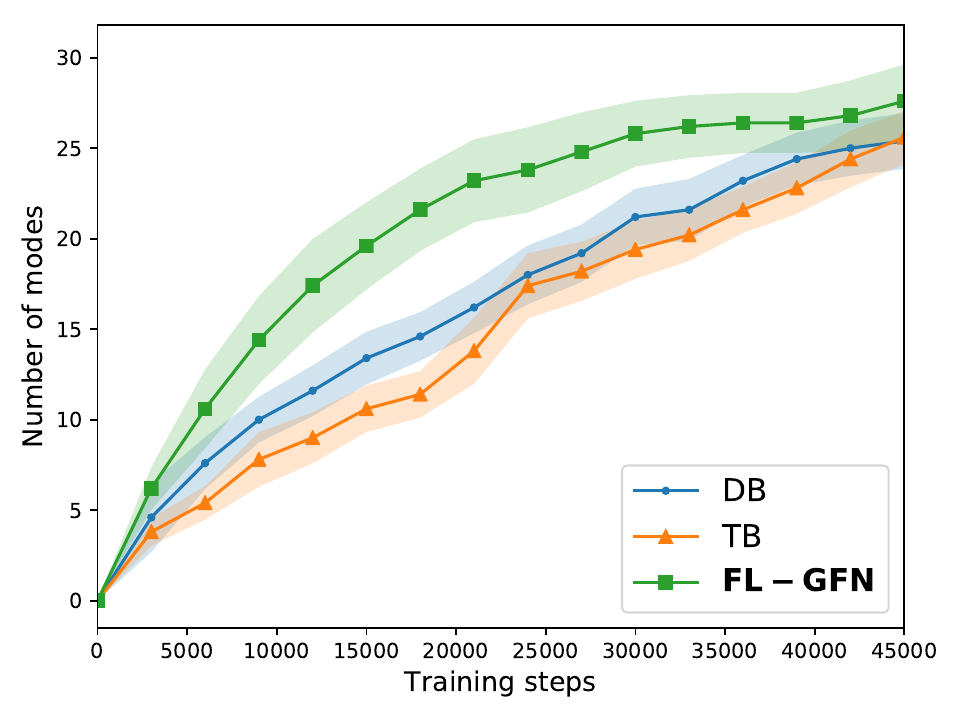}}
\subfloat[]{\includegraphics[width=0.28\linewidth]{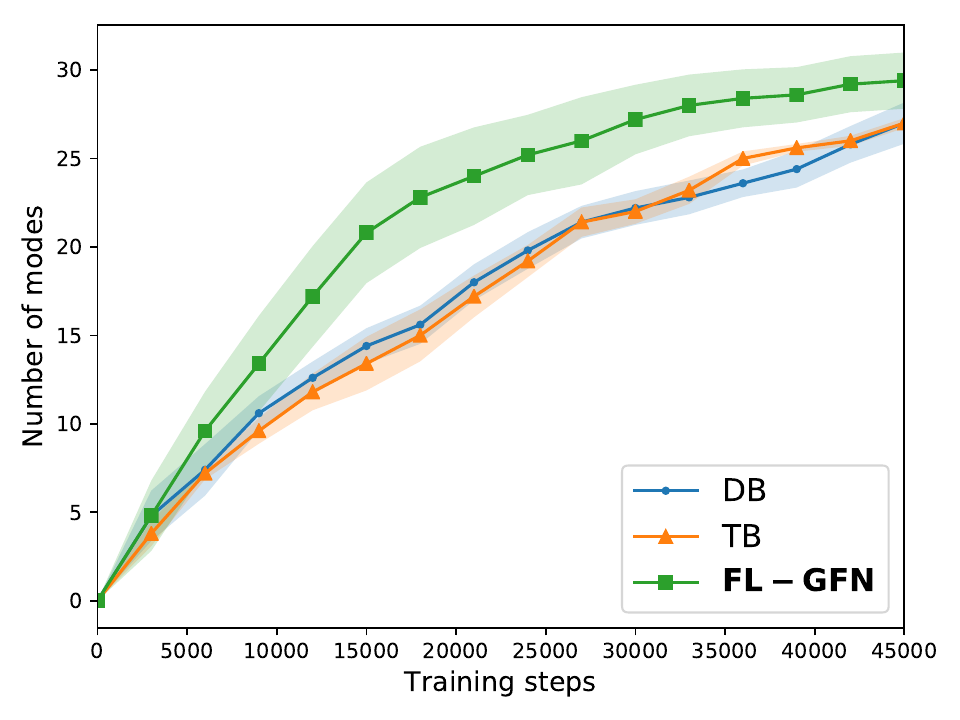}}
\subfloat[]{\includegraphics[width=0.28\linewidth]{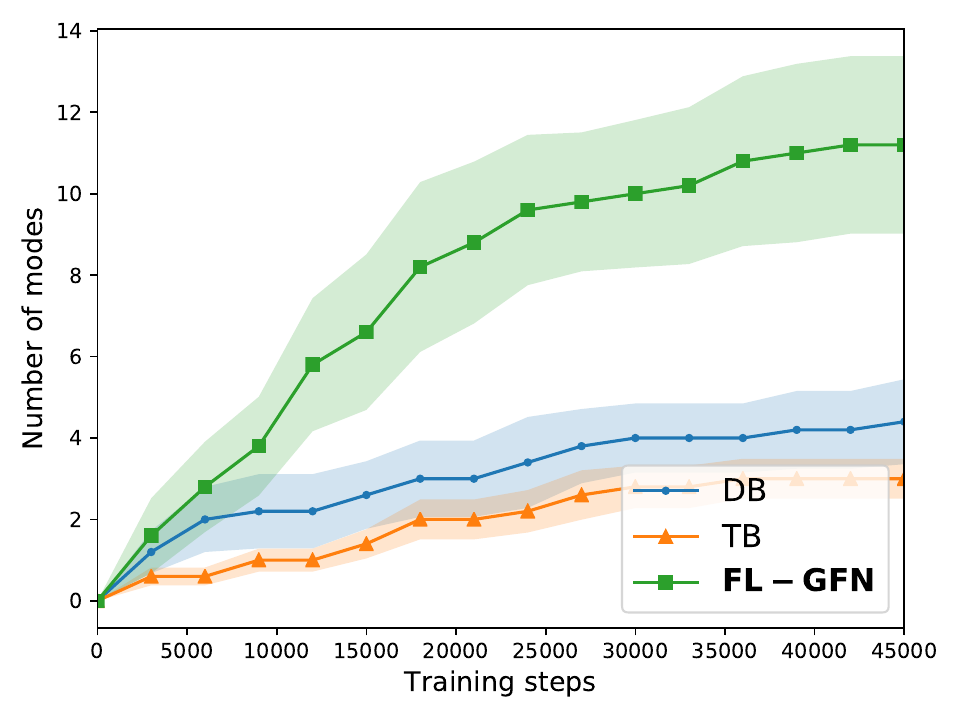}}
\caption{The number of modes discovered by each method in the bit sequence generation task with increasing lengths of the sequences. Different problem sizes are presented from left to right: (a) normal (b) long (c) very long. The advantage of FL-GFN increases with the size of the problem.
}
\label{fig:bit_res}
\vspace*{-1mm}
\end{figure*}

\begin{figure*}[t]
\centering
\vspace*{-2mm}
\subfloat[]{\includegraphics[width=0.25\linewidth]{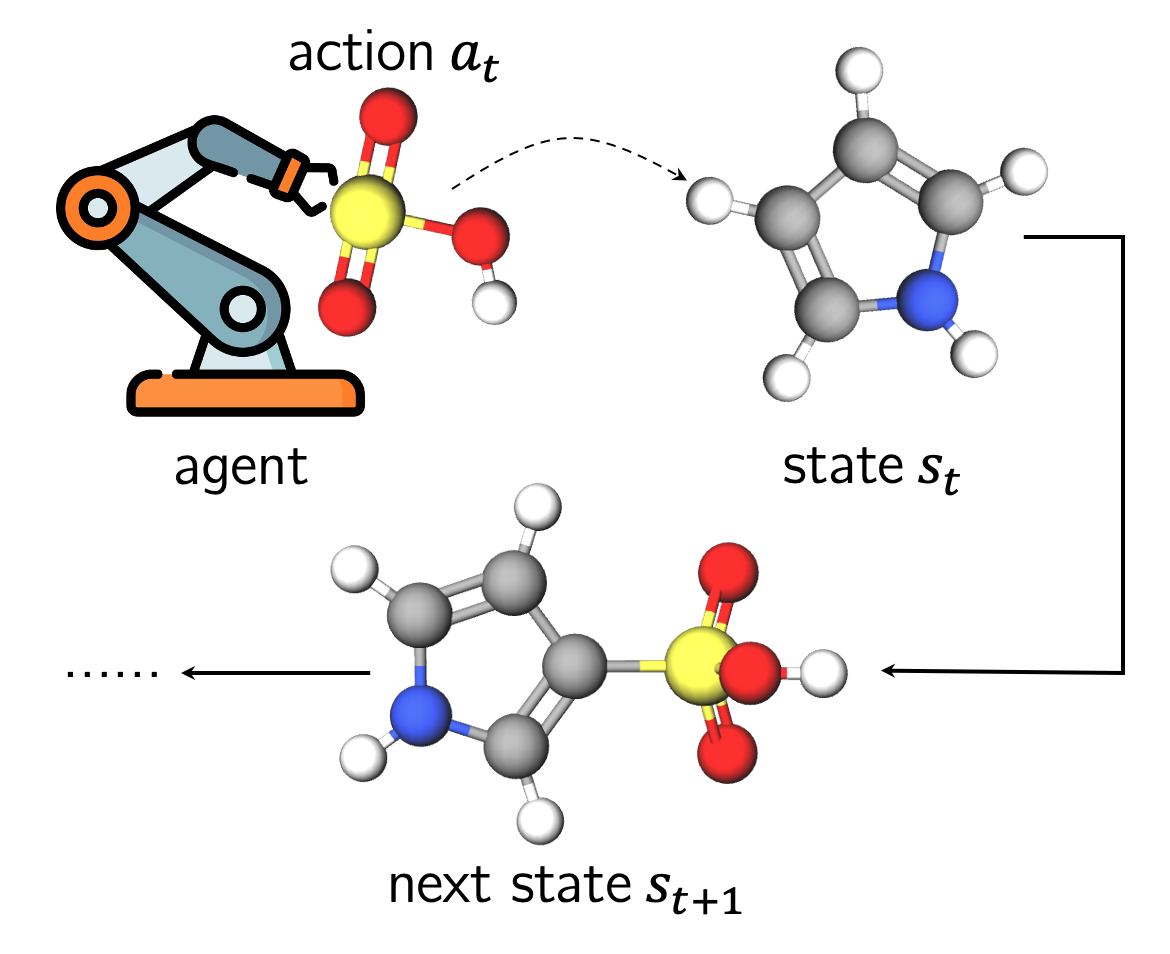}}
\hspace{1mm}
\subfloat[]{\includegraphics[width=0.25\linewidth]{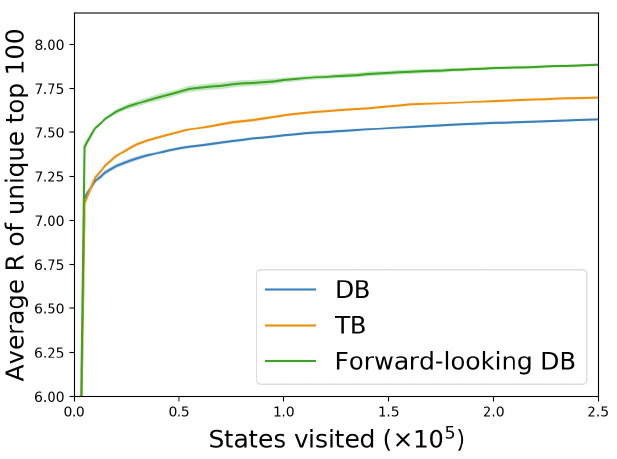}}
\hspace{1mm}
\subfloat[]{\includegraphics[width=0.25\linewidth]{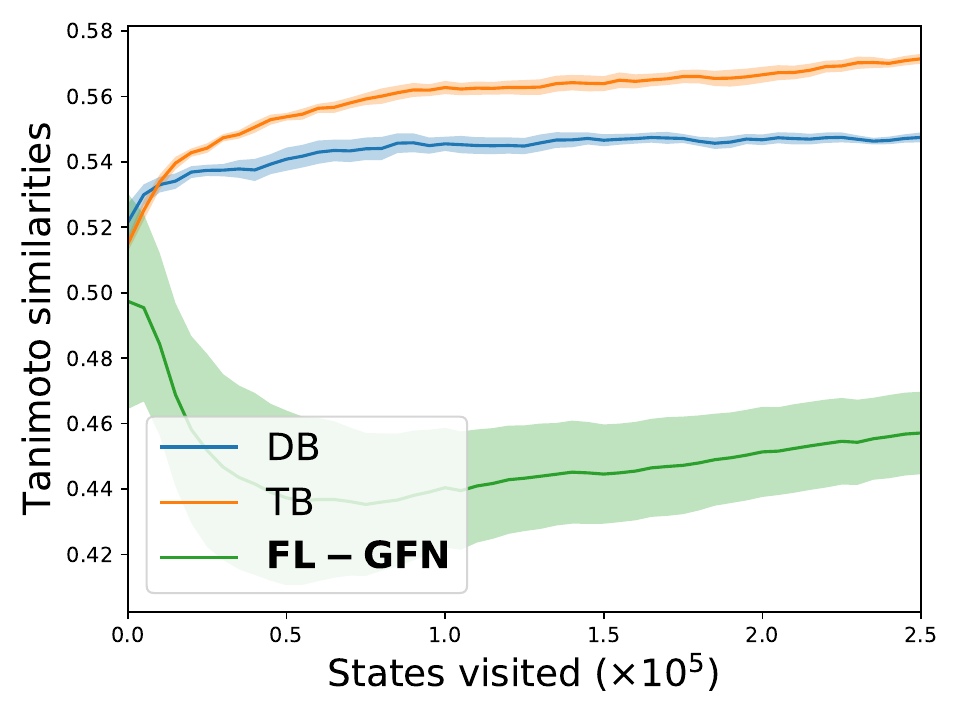}}
\hspace{1mm}
\subfloat[]{\includegraphics[width=0.185\linewidth]{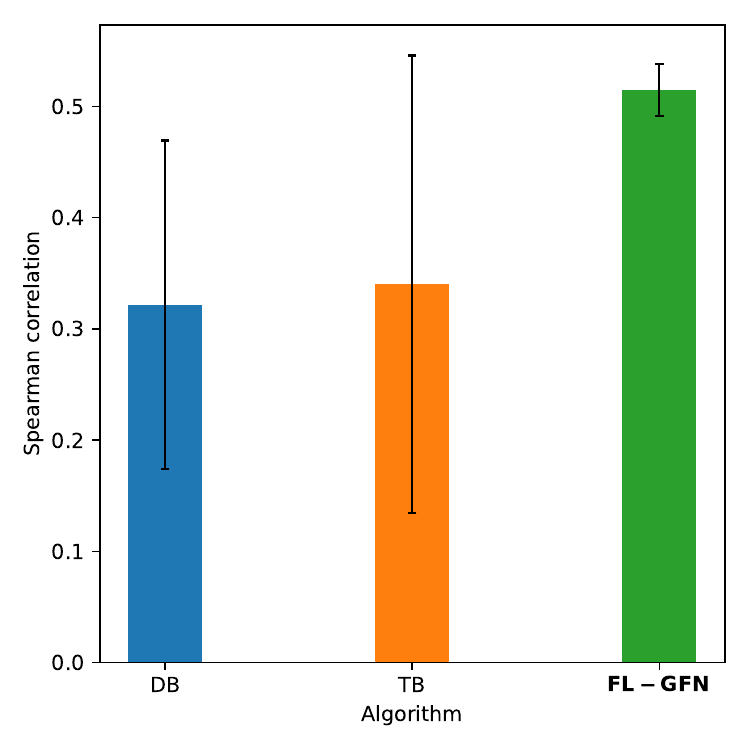}}
\caption{Results on molecule generation task. 
(a) Illustration of the GFlowNet generation pipeline. 
(b) Average reward of the top-$100$ molecule candidates, showing faster and better training of FL-GFN. 
(c) The Tanimoto similarity which quantifies diversity (lower is better), showing greater diversity with FL-GFN.
(d) The correlation between model log likelihood and the log rewards computed on a held-out set, larger with FL-GFN.
}
\label{fig:mol_res}
\vspace*{-3mm}
\end{figure*}

\subsubsection{Learning with Incomplete Trajectories}
We now investigate the ability of FL-GFN when given only incomplete trajectories, without access to terminal states. 
Here, the length of incomplete trajectories is uniformly distributed in $[1, |S|-1]$, with $|S|$ the length of complete trajectories.
We compare the performance of FL-GFN trained with incomplete trajectories only with its counterpart that is learned with complete trajectories.
We include ordinary DB and SubTB approaches in the comparison for completeness (with the ``partial'' tag in the figure), although they cannot achieve informative learning signals without access to terminal states and rewards (and the apparent improvements in the small set generation in Figure~\ref{fig:set_partial}(a) are only due to randomly sampling more trajectories and finding some good ones).
Note that TB cannot be implemented with incomplete trajectories, as the learning objective of TB involves the terminal reward.

Figure~\ref{fig:set_partial} shows the average reward of unique top-$100$ sets discovered by each method when learned with incomplete trajectories, where the x-axis corresponds to the number of state transitions (interactions with the environment). Unlike the earlier training methods, FL-GFN (partial) is able to learn well at different scales. 
More interestingly, it performs closely to its counterpart trained with full trajectories, validating the ability to learn from only incomplete trajectories.

\subsection{Bit Sequence Generation}
We consider the bit sequence generation task from~\citet{malkin2022trajectory}.
At each time step, the agent chooses to append a $k$-bit ``word" (with $k=4$).
The reward function has modes at a predefined fixed set $M$ of sequences~\citep{malkin2022trajectory}: ${\cal E}(x)=\min_{y \in M} d(x, y)$, where $d$ is the edit distance.
Intermediate transition energies for FL-GFN are obtained by applying the above energy function on intermediate states as well.
We consider increasing sequence lengths, as detailed in Appendix~\ref{app:bit_setup}.
We find that FL-GFN learns more efficiently in more complex tasks by comparing it with DB and TB, which are the most competitive baselines in this task as shown by \citet{malkin2022trajectory},
and evaluate the methods in terms of the number of modes discovered.

\paragraph{Results.} Figure~\ref{fig:bit_res} shows the number of modes discovered by each method during training in bit sequence generation with different lengths.
FL-GFN outperforms baselines in learning efficiency and also discovers more modes, with a more significant gap for longer sequences, suggesting that this is due to more efficient credit assignment.

\subsection{Molecule Generation} \label{sec:mol}
We apply FL-GFN to the more challenging and larger-scale molecule generation task~\citep{bengio2021flow,malkin2022trajectory} as illustrated in Figure~\ref{fig:mol_res}(a), with a large state space (about $10^{16}$) and action space (from $100$ to $2000$). 
A molecule is represented by a graph whose nodes are taken from a vocabulary of molecular building blocks.
We want to discover diverse and high-quality molecules with low binding energy to the soluble epoxide hydrolase (sEH) protein, where the binding energy is computed by a pretrained proxy model, which can be applied to intermediate graphs. Actions consist of adding a molecular block at a selected node in a molecular graph, or going to a terminal state to stop generation. 
TB and DB baselines are implemented based on existing GFlowNet open-source code\footnote{\url{https://github.com/GFNOrg/gflownet/tree/trajectory_balance/mols}}.
We report the mean and standard variance for each algorithm with three random seeds as in~\citet{malkin2022trajectory}.
More details of this setup are in Appendix~\ref{app:mol_setup}.

\paragraph{Results.} 
Figure~\ref{fig:mol_res}(b) shows the advantage of FL-GFN in terms of faster and better training, with the average reward of the top-$100$ unique molecules discovered by each method, which evaluates the quality of generated solutions, while Figure~\ref{fig:mol_res}(c) highlights diversity (as measured by the Tanimoto similarities, lower is better), following~\citet{bengio2021flow}.
Visualization of the top-$3$ molecules (and their diversity) discovered by each method in a run is shown in Appendix (Figure~\ref{fig:mol_vis}).
Figure~\ref{fig:mol_res}(d) shows the improvement brought by FL-GFN to the Spearman correlations between the log-reward $\log R(x)$ and the log-sampling probability $\log P_F^\top(x)$ (following the methodology from ~\citet{bengio2021flow}) on a set of test molecules.
The results demonstrate consistent and significant performance improvement and faster training of FL-GFN in the more complex and challenging molecule generation task.

\section{Conclusion}
In this paper, we propose Forward-looking GFlowNets (FL-GFN), a new GFlowNet formulation which exploits the computability of a per-state or per-transition energy function. FL-GFN can be trained to sample proportionally from the target reward distribution, while speeding up training due to its more efficient credit assignment mechanism. It can even be trained given incomplete trajectories when it does not have access to terminal states, which was a requirement for previous GFlowNet learning objectives. We conduct extensive experiments to demonstrate the effectiveness of FL-GFN, which can scale to complex and challenging tasks, such as molecular graph generation.

\section*{Acknowledgments}

The authors acknowledge funding from CIFAR, Genentech, Samsung, and IBM. We would also like to thank Moksh Jain, Edward Hu, Cristian Dragos Manta, and Hadi Nekoei for valuable discussions.

\bibliography{icml-submission}
\bibliographystyle{icml2023}

\clearpage
\appendix
\onecolumn
\section{Proof of Proposition 4.2} \label{app:proof}
{\textbf{Proposition 4.2}}
{\emph{
Suppose that Assumption~\ref{as:general-energy} is satisfied, which makes it possible to define a forward-looking flow as per Eq.~(\ref{eq:fl_flow}). If $\mathcal{L}(s, s') = 0$ for all transitions, then the forward policy $P_F(s'|s; \theta)$ samples proportionally to the reward function.
}}

\begin{proof}
We give a simple proof by reduction to the DB training theorem \citep{bengio2021foundations}, which states that if for some state flow function $F$ and policies $P_F$ and $P_B$ the DB constraint as in Eq.~(\ref{eq:db_constraint}) holds for all transitions $s\rightarrow s'$, then $P_F$ samples proportionally to the reward.

Suppose $\widetilde{F}$, $P_F$, and $P_B$ satisfy Eq.~(\ref{eq:fmconstraint_fldb}). Define a state flow function $\hat F$ by $\hat F(s)\eqdef e^{-\gE(s)}\widetilde{F}(s)$. By direct algebraic manipulation of Eq.~(\ref{eq:fmconstraint_fldb}), we obtain, for all transitions $s\rightarrow s'$,
\begin{align*}
\hat F(s)P_F(s'|s)
&=\frac{e^{-\gE(s)}}{e^{-\gE(s')}}\hat F(s')P_B(s|s')e^{-\gE(s\rightarrow s')}\\
&=\hat F(s')P_B(s|s').
\end{align*}
Therefore, the triple $(\hat F,P_F,P_B)$ satisfies DB, implying that $P_F$ samples proportionally to the reward.
\end{proof}

\section{Experimental Details} 
All baseline methods are implemented based on the open-source implementation as described in the following sections by following the default hyperparameters and setup. The code will be released upon publication of the paper.

\subsection{Set Generation} \label{app:set_setup}
We study the set generation task with increasing scales including small, medium, and large.
The dimension of action space $|U|$ for small, medium, and large is $30$, $80$, and $100$, respectively. The size of the target set to generate $|S|$ is $20$, $60$, $80$, respectively.
The predefined energy $\mathcal{E}(t)$ for each element $t$ is randomly generated in the range $[-1, 1]$, and $|S|/10$ of the elements have the same energy (resulting in multiple optimal solutions).
We implement DB, TB, SubTB and FL-DB based on the open-source codes\footnote{\url{https://github.com/GFNOrg/gflownet}}.
The GFlowNet model is a feedforward network that consists of $2$ hidden layers with $256$ hidden units per layer which uses the LeakyReLU activation function.
We sample a parallel of $16$ rollouts from the environment for training all of the models.
The GFlowNet model is trained based on the Adam~\citep{kingma2014adam} optimizer with a learning rate of $0.001$ for DB, SubTB, and FL-DB, where we use a larger learning rate of $0.1$ for the learnable parameter $Z$ for TB following~\citep{malkin2022trajectory}.

\subsection{Bit Sequence Generation} \label{app:bit_setup}
We implement all baselines based on \citet{malkin2022trajectory} and follow the default hyperparameters and setup as in \citet{malkin2022trajectory}. 
We study the bit sequence generation tasks with increasing sequence lengths $n$ including normal ($n=120$), long ($n=140$), and very long ($n=160$).
The GFlowNet model is a Transformer~\citep{vaswani2017attention} consisting of $3$ hidden layers with $64$ hidden units per layer and has $8$ attention heads. 
The size of the minibatch is $16$, and the random action probability is set to $0.0005$ for performing $\epsilon$-greedy exploration. The reward exponent is set to $3$, and we use a sampling temperature of $1$ for the forward policy $P_F$ for the GFlowNet models. 
The learning rate for the policy parameters is $1\times 10^{-4}$ for TB, and the learning rate for the learnable parameter $Z$ is $1 \times 10^{-3}$.
The learning rate is $5 \times 10^{-3}$ for the DB and FL-GFN variants.

\clearpage
\subsection{Molecule Discovery} \label{app:mol_setup}
All baselines are implemented based on the open-source codes\footnote{\url{https://github.com/GFNOrg/gflownet/tree/master/mols}}.
We use a reward proxy provided in \citet{bengio2021flow}. 
We use Message Passing Neural Networks (MPNN) for the network architecture for all GFlowNet models, as the molecule is represented as an atom graph. We follow the default hyperparameters and setup as in \citet{malkin2022trajectory}. We fine-tune the reward exponent by grid search following~\citep{malkin2022trajectory} and set it to be $4$. We use a random action probability of $0.1$ for performing $\epsilon$-greedy exploration, and the learning rate is $5 \times 10^{-4}$.

\begin{figure}[!h]
\centering
\subfloat[Detailed balance (DB)]{\includegraphics[width=0.7\linewidth]{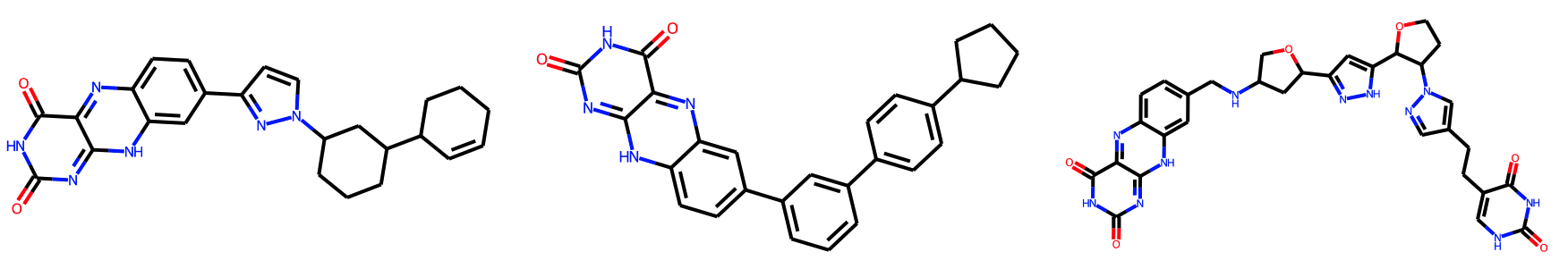}}\\
\subfloat[Trajectory balance (TB)]{\includegraphics[width=0.7\linewidth]{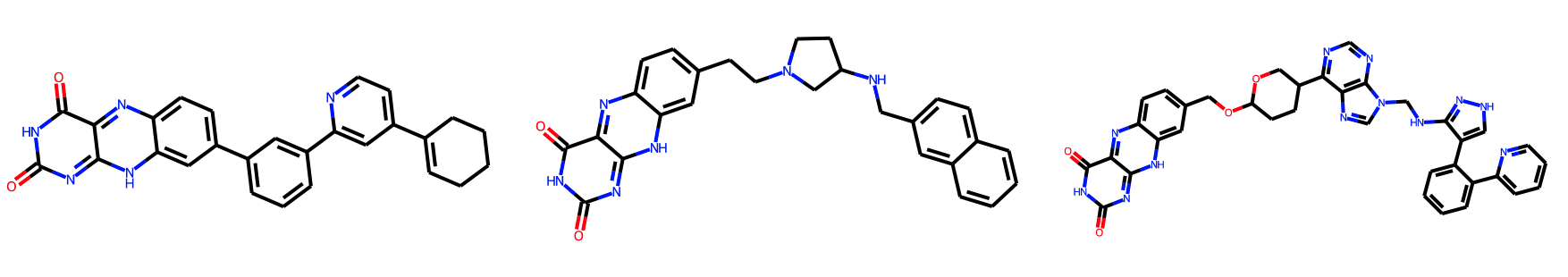}}\\
\subfloat[FL-GFN]{\includegraphics[width=0.7\linewidth]{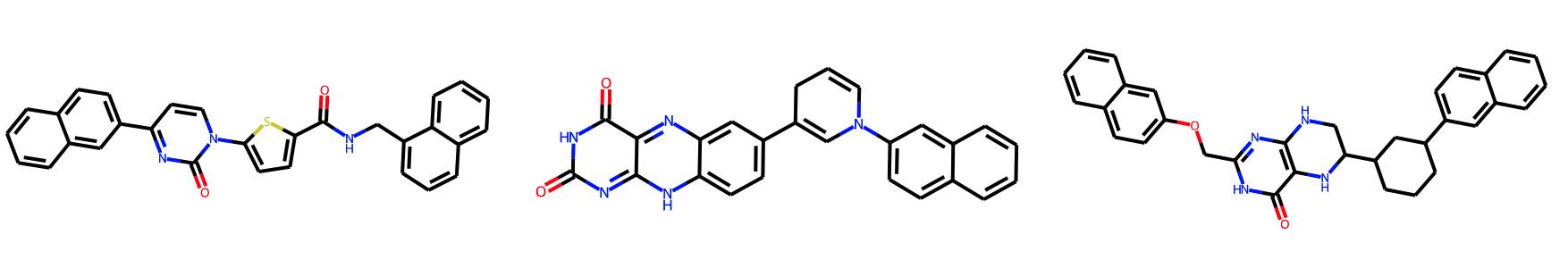}} 
\caption{Visualization of top-$3$ molecules generated by different methods.}
\label{fig:mol_vis}
\end{figure}

\end{document}

%% file: math_commands.tex
\usepackage{amsmath,amsfonts,bm}

\def\eqref#1{eq.~\ref{#1}}

\def\1{\bm{1}}

\DeclareMathAlphabet{\mathsfit}{\encodingdefault}{\sfdefault}{m}{sl}
\SetMathAlphabet{\mathsfit}{bold}{\encodingdefault}{\sfdefault}{bx}{n}

\def\gE{{\mathcal{E}}}

\def\gS{{\mathcal{S}}}

\def\gX{{\mathcal{X}}}

\newcommand{\R}{\mathbb{R}}



%% file: icml-submission.bbl
\begin{thebibliography}{48}
\providecommand{\natexlab}[1]{#1}
\providecommand{\url}[1]{\texttt{#1}}
\expandafter\ifx\csname urlstyle\endcsname\relax
  \providecommand{\doi}[1]{doi: #1}\else
  \providecommand{\doi}{doi: \begingroup \urlstyle{rm}\Url}\fi

\bibitem[Andrieu et~al.(2003)Andrieu, De~Freitas, Doucet, and
  Jordan]{andrieu2003introduction}
Andrieu, C., De~Freitas, N., Doucet, A., and Jordan, M.~I.
\newblock An introduction to mcmc for machine learning.
\newblock \emph{Machine learning}, 50\penalty0 (1):\penalty0 5--43, 2003.

\bibitem[Baars(1993)]{baars1993cognitive}
Baars, B.~J.
\newblock \emph{A cognitive theory of consciousness}.
\newblock Cambridge University Press, 1993.

\bibitem[Baddeley(1992)]{baddeley1992working}
Baddeley, A.
\newblock Working memory.
\newblock \emph{Science}, 255\penalty0 (5044):\penalty0 556--559, 1992.

\bibitem[Bengio et~al.(2021{\natexlab{a}})Bengio, Jain, Korablyov, Precup, and
  Bengio]{bengio2021flow}
Bengio, E., Jain, M., Korablyov, M., Precup, D., and Bengio, Y.
\newblock Flow network based generative models for non-iterative diverse
  candidate generation.
\newblock \emph{Neural Information Processing Systems (NeurIPS)},
  2021{\natexlab{a}}.

\bibitem[Bengio et~al.(2013)Bengio, Mesnil, Dauphin, and
  Rifai]{bengio2013better}
Bengio, Y., Mesnil, G., Dauphin, Y., and Rifai, S.
\newblock Better mixing via deep representations.
\newblock \emph{International Conference on Machine Learning (ICML)}, 2013.

\bibitem[Bengio et~al.(2021{\natexlab{b}})Bengio, Lahlou, Deleu, Hu, Tiwari,
  and Bengio]{bengio2021foundations}
Bengio, Y., Lahlou, S., Deleu, T., Hu, E., Tiwari, M., and Bengio, E.
\newblock {GFlowNet} foundations.
\newblock \emph{arXiv preprint 2111.09266}, 2021{\natexlab{b}}.

\bibitem[Bengio et~al.(2022)Bengio, Malkin, and Jain]{gfn-tutorial2022}
Bengio, Y., Malkin, N., and Jain, M.
\newblock The {GFlowNet} {T}utorial.
\newblock
  https://milayb.notion.site/The-GFlowNet-Tutorial-95434ef0e2d94c24aab90e69b30be9b3,
  2022.

\bibitem[Cowan(1999)]{cowan1999embedded}
Cowan, N.
\newblock An embedded-processes model of working memory.
\newblock 1999.

\bibitem[Dehaene et~al.(1998)Dehaene, Kerszberg, and
  Changeux]{dehaene1998neuronal}
Dehaene, S., Kerszberg, M., and Changeux, J.-P.
\newblock A neuronal model of a global workspace in effortful cognitive tasks.
\newblock \emph{Proceedings of the national Academy of Sciences}, 95\penalty0
  (24):\penalty0 14529--14534, 1998.

\bibitem[Dehaene et~al.(2017)Dehaene, Lau, and
  Kouider]{dehaene2017consciousness}
Dehaene, S., Lau, H., and Kouider, S.
\newblock What is consciousness, and could machines have it?
\newblock \emph{Science}, 358\penalty0 (6362):\penalty0 486--492, 2017.

\bibitem[Deleu et~al.(2022)Deleu, G\'{o}is, Emezue, Rankawat, Lacoste-Julien,
  Bauer, and Bengio]{deleu2022bayesian}
Deleu, T., G\'{o}is, A., Emezue, C., Rankawat, M., Lacoste-Julien, S., Bauer,
  S., and Bengio, Y.
\newblock Bayesian structure learning with generative flow networks.
\newblock \emph{Uncertainty in Artificial Intelligence (UAI)}, 2022.

\bibitem[Fujimoto et~al.(2018)Fujimoto, Hoof, and
  Meger]{fujimoto2018addressing}
Fujimoto, S., Hoof, H., and Meger, D.
\newblock Addressing function approximation error in actor-critic methods.
\newblock In \emph{International conference on machine learning}, pp.\
  1587--1596. PMLR, 2018.

\bibitem[Goodfellow et~al.(2014)Goodfellow, Pouget-Abadie, Mirza, Xu,
  Warde-Farley, Ozair, Courville, and Bengio]{goodfellow2014generative}
Goodfellow, I., Pouget-Abadie, J., Mirza, M., Xu, B., Warde-Farley, D., Ozair,
  S., Courville, A., and Bengio, Y.
\newblock Generative adversarial nets.
\newblock \emph{Neural Information Processing Systems (NIPS)}, pp.\
  2672--2680, 2014.

\bibitem[Goodfellow et~al.(2016)Goodfellow, Bengio, and
  Courville]{goodfellow2016deep}
Goodfellow, I., Bengio, Y., and Courville, A.
\newblock \emph{Deep learning}.
\newblock MIT press, 2016.

\bibitem[Haarnoja et~al.(2017)Haarnoja, Tang, Abbeel, and
  Levine]{haarnoja2017reinforcement}
Haarnoja, T., Tang, H., Abbeel, P., and Levine, S.
\newblock Reinforcement learning with deep energy-based policies.
\newblock \emph{International Conference on Machine Learning (ICML)}, 2017.

\bibitem[Haarnoja et~al.(2018)Haarnoja, Zhou, Abbeel, and
  Levine]{haarnoja2018soft}
Haarnoja, T., Zhou, A., Abbeel, P., and Levine, S.
\newblock Soft actor-critic: Off-policy maximum entropy deep reinforcement
  learning with a stochastic actor.
\newblock \emph{International Conference on Machine Learning (ICML)}, 2018.

\bibitem[Hastings(1970)]{hastings1970monte}
Hastings, W.~K.
\newblock Monte carlo sampling methods using markov chains and their
  applications.
\newblock 1970.

\bibitem[Ho et~al.(2020)Ho, Jain, and Abbeel]{ho2020denoising}
Ho, J., Jain, A., and Abbeel, P.
\newblock Denoising diffusion probabilistic models.
\newblock \emph{Advances in Neural Information Processing Systems},
  33:\penalty0 6840--6851, 2020.

\bibitem[Hu et~al.(2023)Hu, Malkin, Jain, Everett, Graikos, and Bengio]{gfn-em}
Hu, E.~J., Malkin, N., Jain, M., Everett, K., Graikos, A., and Bengio, Y.
\newblock {GFlowNet-EM} for learning compositional latent variable models.
\newblock \emph{International Conference on Machine Learning (ICML)}, 2023.

\bibitem[Jain et~al.(2022{\natexlab{a}})Jain, Bengio, Hernandez-Garcia,
  Rector-Brooks, Dossou, Ekbote, Fu, Zhang, Kilgour, Zhang, Simine, Das, and
  Bengio]{jain2022biological}
Jain, M., Bengio, E., Hernandez-Garcia, A., Rector-Brooks, J., Dossou, B.~F.,
  Ekbote, C., Fu, J., Zhang, T., Kilgour, M., Zhang, D., Simine, L., Das, P.,
  and Bengio, Y.
\newblock Biological sequence design with {GFlowNets}.
\newblock \emph{International Conference on Machine Learning (ICML)},
  2022{\natexlab{a}}.

\bibitem[Jain et~al.(2022{\natexlab{b}})Jain, Raparthy, Hernandez-Garcia,
  Rector-Brooks, Bengio, Miret, and Bengio]{jain2022multi}
Jain, M., Raparthy, S.~C., Hernandez-Garcia, A., Rector-Brooks, J., Bengio, Y.,
  Miret, S., and Bengio, E.
\newblock Multi-objective gflownets.
\newblock \emph{arXiv preprint arXiv:2210.12765}, 2022{\natexlab{b}}.

\bibitem[Kingma \& Ba(2015)Kingma and Ba]{kingma2014adam}
Kingma, D.~P. and Ba, J.
\newblock Adam: A method for stochastic optimization.
\newblock \emph{International Conference on Learning Representations (ICLR)},
  2015.

\bibitem[Kingma \& Welling(2013)Kingma and Welling]{kingma2013auto}
Kingma, D.~P. and Welling, M.
\newblock Auto-encoding variational bayes.
\newblock \emph{arXiv preprint arXiv:1312.6114}, 2013.

\bibitem[Kingma \& Welling(2014)Kingma and Welling]{Kingma2014AutoEncodingVB}
Kingma, D.~P. and Welling, M.
\newblock Auto-encoding variational {Bayes}.
\newblock \emph{International Conference on Learning Representations (ICLR)},
  2014.

\bibitem[Lahlou et~al.(2023)Lahlou, Deleu, Lemos, Zhang, Volokhova,
  Hern\'{a}ndez-Garc\'{i}a, Ezzine, Bengio, and
  Malkin]{lahlou2023continuousgfn}
Lahlou, S., Deleu, T., Lemos, P., Zhang, D., Volokhova, A.,
  Hern\'{a}ndez-Garc\'{i}a, A., Ezzine, L.~N., Bengio, Y., and Malkin, N.
\newblock A theory of continuous generative flow networks.
\newblock \emph{International Conference on Machine Learning (ICML)}, 2023.

\bibitem[Lillicrap et~al.(2015)Lillicrap, Hunt, Pritzel, Heess, Erez, Tassa,
  Silver, and Wierstra]{lillicrap2015continuous}
Lillicrap, T.~P., Hunt, J.~J., Pritzel, A., Heess, N., Erez, T., Tassa, Y.,
  Silver, D., and Wierstra, D.
\newblock Continuous control with deep reinforcement learning.
\newblock \emph{arXiv preprint arXiv:1509.02971}, 2015.

\bibitem[Madan et~al.(2022)Madan, Rector-Brooks, Korablyov, Bengio, Jain, Nica,
  Bosc, Bengio, and Malkin]{madan2022learning}
Madan, K., Rector-Brooks, J., Korablyov, M., Bengio, E., Jain, M., Nica, A.,
  Bosc, T., Bengio, Y., and Malkin, N.
\newblock Learning {GFlowNets} from partial episodes for improved convergence
  and stability.
\newblock \emph{arXiv preprint 2209.12782}, 2022.

\bibitem[Malkin et~al.(2022{\natexlab{a}})Malkin, Jain, Bengio, Sun, and
  Bengio]{malkin2022trajectory}
Malkin, N., Jain, M., Bengio, E., Sun, C., and Bengio, Y.
\newblock Trajectory balance: Improved credit assignment in {GFlowNets}.
\newblock \emph{Neural Information Processing Systems (NeurIPS)},
  2022{\natexlab{a}}.

\bibitem[Malkin et~al.(2022{\natexlab{b}})Malkin, Lahlou, Deleu, Ji, Hu,
  Everett, Zhang, and Bengio]{malkin2022gfnhvi}
Malkin, N., Lahlou, S., Deleu, T., Ji, X., Hu, E., Everett, K., Zhang, D., and
  Bengio, Y.
\newblock Gflownets and variational inference.
\newblock \emph{arXiv preprint arXiv:2210.00580}, 2022{\natexlab{b}}.

\bibitem[Metropolis et~al.(1953)Metropolis, Rosenbluth, Rosenbluth, Teller, and
  Teller]{metropolis1953equation}
Metropolis, N., Rosenbluth, A.~W., Rosenbluth, M.~N., Teller, A.~H., and
  Teller, E.
\newblock Equation of state calculations by fast computing machines.
\newblock \emph{The journal of chemical physics}, 21\penalty0 (6):\penalty0
  1087--1092, 1953.

\bibitem[Mnih et~al.(2015)Mnih, Kavukcuoglu, Silver, Rusu, Veness, Bellemare,
  Graves, Riedmiller, Fidjeland, Ostrovski, et~al.]{mnih2015human}
Mnih, V., Kavukcuoglu, K., Silver, D., Rusu, A.~A., Veness, J., Bellemare,
  M.~G., Graves, A., Riedmiller, M., Fidjeland, A.~K., Ostrovski, G., et~al.
\newblock Human-level control through deep reinforcement learning.
\newblock \emph{nature}, 518\penalty0 (7540):\penalty0 529--533, 2015.

\bibitem[Nishikawa-Toomey et~al.(2022)Nishikawa-Toomey, Deleu, Subramanian,
  Bengio, and Charlin]{nishikawa2022bayesian}
Nishikawa-Toomey, M., Deleu, T., Subramanian, J., Bengio, Y., and Charlin, L.
\newblock Bayesian learning of causal structure and mechanisms with {GFlowNets}
  and variational bayes.
\newblock \emph{arXiv preprint 2211.02763}, 2022.

\bibitem[Pan et~al.(2022)Pan, Zhang, Courville, Huang, and Bengio]{pan2022gafn}
Pan, L., Zhang, D., Courville, A., Huang, L., and Bengio, Y.
\newblock Generative augmented flow networks.
\newblock \emph{arXiv preprint 2210.03308}, 2022.

\bibitem[Pan et~al.(2023)Pan, Zhang, Jain, Huang, and
  Bengio]{pan2023stochastic}
Pan, L., Zhang, D., Jain, M., Huang, L., and Bengio, Y.
\newblock Stochastic generative flow networks.
\newblock \emph{Uncertainty in Artificial Intelligence (UAI)}, 2023.

\bibitem[Ren et~al.(2022)Ren, Guo, Zhou, and Peng]{ren2021learning}
Ren, Z., Guo, R., Zhou, Y., and Peng, J.
\newblock Learning long-term reward redistribution via randomized return
  decomposition.
\newblock \emph{International Conference on Learning Representations (ICLR)},
  2022.

\bibitem[Rezende et~al.(2014)Rezende, Mohamed, and
  Wierstra]{JimenezRezende2014StochasticBA}
Rezende, D.~J., Mohamed, S., and Wierstra, D.
\newblock Stochastic backpropagation and approximate inference in deep
  generative models.
\newblock \emph{International Conference on Machine Learning (ICML)}, 2014.

\bibitem[Salakhutdinov(2009)]{salakhutdinov2009learning}
Salakhutdinov, R.
\newblock Learning in markov random fields using tempered transitions.
\newblock \emph{Neural Information Processing Systems (NIPS)}, 2009.

\bibitem[Shanahan(2006)]{shanahan2006cognitive}
Shanahan, M.
\newblock A cognitive architecture that combines internal simulation with a
  global workspace.
\newblock \emph{Consciousness and cognition}, 15\penalty0 (2):\penalty0
  433--449, 2006.

\bibitem[Shanahan(2010)]{shanahan2010embodiment}
Shanahan, M.
\newblock \emph{Embodiment and the inner life: Cognition and Consciousness in
  the Space of Possible Minds}.
\newblock Oxford University Press, USA, 2010.

\bibitem[Shanahan(2012)]{shanahan2012brain}
Shanahan, M.
\newblock The brain's connective core and its role in animal cognition.
\newblock \emph{Philosophical Transactions of the Royal Society B: Biological
  Sciences}, 367\penalty0 (1603):\penalty0 2704--2714, 2012.

\bibitem[Shanahan \& Baars(2005)Shanahan and Baars]{shanahan2005applying}
Shanahan, M. and Baars, B.
\newblock Applying global workspace theory to the frame problem.
\newblock \emph{Cognition}, 98\penalty0 (2):\penalty0 157--176, 2005.

\bibitem[Sutton \& Barto(2018)Sutton and Barto]{sutton2018reinforcement}
Sutton, R.~S. and Barto, A.~G.
\newblock \emph{Reinforcement learning: An introduction}.
\newblock MIT Press, 2018.

\bibitem[Vaswani et~al.(2017)Vaswani, Shazeer, Parmar, Uszkoreit, Jones, Gomez,
  Kaiser, and Polosukhin]{vaswani2017attention}
Vaswani, A., Shazeer, N., Parmar, N., Uszkoreit, J., Jones, L., Gomez, A.~N.,
  Kaiser, {\L}., and Polosukhin, I.
\newblock Attention is all you need.
\newblock \emph{Neural Information Processing Systems (NIPS)}, 2017.

\bibitem[Zhang et~al.(2022{\natexlab{a}})Zhang, Chen, Malkin, and
  Bengio]{zhang2022unifying}
Zhang, D., Chen, R. T.~Q., Malkin, N., and Bengio, Y.
\newblock Unifying generative models with {GFlowNets}.
\newblock \emph{arXiv preprint 2209.02606}, 2022{\natexlab{a}}.

\bibitem[Zhang et~al.(2022{\natexlab{b}})Zhang, Courville, Bengio, Zheng,
  Zhang, and Chen]{Zhang2022LatentSM}
Zhang, D., Courville, A.~C., Bengio, Y., Zheng, Q., Zhang, A., and Chen, R.
  T.~Q.
\newblock Latent state marginalization as a low-cost approach for improving
  exploration.
\newblock \emph{ArXiv}, abs/2210.00999, 2022{\natexlab{b}}.

\bibitem[Zhang et~al.(2022{\natexlab{c}})Zhang, Malkin, Liu, Volokhova,
  Courville, and Bengio]{zhang2022generative}
Zhang, D., Malkin, N., Liu, Z., Volokhova, A., Courville, A., and Bengio, Y.
\newblock Generative flow networks for discrete probabilistic modeling.
\newblock \emph{International Conference on Machine Learning (ICML)},
  2022{\natexlab{c}}.

\bibitem[Zhang et~al.(2023)Zhang, Pan, Chen, Courville, and
  Bengio]{zhang2023distributional}
Zhang, D., Pan, L., Chen, R.~T., Courville, A., and Bengio, Y.
\newblock Distributional gflownets with quantile flows.
\newblock \emph{arXiv preprint arXiv:2302.05793}, 2023.

\bibitem[Zimmermann et~al.(2022)Zimmermann, Lindsten, van~de Meent, and
  Naesseth]{zimmermann2022variational}
Zimmermann, H., Lindsten, F., van~de Meent, J.-W., and Naesseth, C.~A.
\newblock A variational perspective on generative flow networks.
\newblock \emph{arXiv preprint 2210.07992}, 2022.

\end{thebibliography}
